\documentclass{article}

\usepackage{arxiv}

\usepackage[utf8]{inputenc} % allow utf-8 input
\usepackage[T1]{fontenc}    % use 8-bit T1 fonts
\usepackage{hyperref}       % hyperlinks
\usepackage{url}            % simple URL typesetting
\usepackage{booktabs}       % professional-quality tables
\usepackage{amsfonts}       % blackboard math symbols
\usepackage{nicefrac}       % compact symbols for 1/2, etc.
\usepackage{microtype}      % microtypography
\usepackage{lipsum}
\usepackage{amsthm}
\usepackage{amssymb}
\usepackage{amsmath,empheq, amsfonts}
\usepackage{url}
\usepackage{bm}
\usepackage{placeins}
\usepackage{algorithm}
\usepackage{algpseudocode}

\usepackage{booktabs}       % professional-quality tables
\usepackage{subcaption}  
\usepackage{multirow}

\title{Leverage Score Sampling for Complete Mode Coverage in Generative Adversarial Networks}

  \author{Joachim Schreurs \\
  Department of Electrical Engineering\\
  ESAT-STADIUS, KU Leuven\\
  Kasteelpark Arenberg 10, B-3001 Leuven, Belgium \\
  \texttt{joachim.schreurs@kuleuven.be} \And Hannes De Meulemeester \\
  Department of Electrical Engineering\\
  ESAT-STADIUS, KU Leuven\\
  Kasteelpark Arenberg 10, B-3001 Leuven, Belgium\\
  \texttt{hannes.demeulemeester@kuleuven.be}  \And \\
  \And  Micha\"el Fanuel\thanks{Most of this work was done when MF was at KU Leuven.} \\
  UMR 9189 – CRIStAL \\
  Univ. Lille, CNRS, Centrale Lille \\
  F-59000 Lille, France \\
  \texttt{michael.fanuel@univ-lille.fr}\And Bart De Moor  \\
  Department of Electrical Engineering\\
  ESAT-STADIUS, KU Leuven\\
  Kasteelpark Arenberg 10, B-3001 Leuven, Belgium\\
  \texttt{bart.demoor@kuleuven.be}\And Johan A.K. Suykens \\
  Department of Electrical Engineering\\
  ESAT-STADIUS, KU Leuven\\
  Kasteelpark Arenberg 10, B-3001 Leuven, Belgium\\
  \texttt{johan.suykens@kuleuven.be}}

\begin{document}
\maketitle

\begin{abstract}
Commonly, machine learning models minimize an empirical expectation. As a result, the trained models typically perform well for the majority of the data but the performance may deteriorate in less dense regions of the dataset. This issue also arises in generative modeling. A  generative model may overlook underrepresented modes that are less frequent in the empirical data distribution. This problem is known as complete mode coverage. We propose a sampling procedure based on ridge leverage scores which significantly improves mode coverage when compared to standard methods and can easily be combined with any GAN. Ridge leverage scores are computed by using an explicit feature map, associated with the next-to-last layer of a GAN discriminator or of a pre-trained network, or by using an implicit feature map corresponding to a Gaussian kernel. Multiple evaluations against recent approaches of complete mode coverage show a clear improvement when using the proposed sampling strategy.
\end{abstract}

\section{Introduction}

\begin{figure}[h!]
	\centering
    \includegraphics[width=0.6\linewidth]{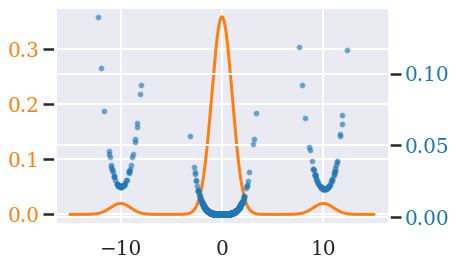}
    \caption{Probability Density Function (orange) and RLS of a sample of this PDF (blue). We take the motivating example from~\cite{zhong2019rethinking}, which consists of a 1D target PDF $p$  with 1 majority mode and 2 minority modes: $p = 0.9 \cdot \mathcal{N}(0,1) + 0.05 \cdot \mathcal{N}(10,1)+0.05 \cdot \mathcal{N}(-10,1)$ given in orange. The RLS distribution is calculated using a Gaussian kernel with $\sigma = 3$ and $\gamma = 10^{-3}$. When sampling a mini-batch from the PDF $p$,  the side modes can be missed. This is resolved by sampling from the RLS distribution.}
	\label{fig:first example}
\end{figure}

Complete mode coverage is a problem of generative models which has been clearly defined and studied in~\cite{zhong2019rethinking}.  In layman's terms, a mode is defined as a local maximum of the data probability density. 
A closely related problem is mode collapse in GANs~\cite{goodfellow2014generative}, which happens when a generative model is only capable of generating samples from a subset of all the modes. 
 Multiple GAN variants have been proposed as a solution to this problem, however proposed solutions often assume that every mode has an (almost)  equal probability of being sampled, which is often not the case in realistic datasets. Regularly, in critical applications, datasets contain a mixture of different subpopulations where the frequency of each subpopulation can be vastly different. 
 The role of less abundant subpopulations in machine learning data has been discussed recently in~\cite{AShortTaleaboutaLongTail}. Also, it is often common to presume that an algorithm does not know the abundance of subpopulations. It is however important that a machine learning model performs well on all subpopulations. A standard example is medical data where some rare diseases are less abundant than common diseases. To illustrate the approach presented in this paper, a motivating example containing one majority mode and two minority modes is given in Figure~\ref{fig:first example}. When sampling a mini-batch from the Probability Density Function (PDF) $p$, the side modes can be missed. We observe empirically that this is resolved by sampling from the ridge leverage score (RLS) distribution (see Section~\ref{sec:RLS_Sampling}), which has been extensively used in randomized linear algebra and kernel methods. Figure~\ref{fig:first example} shows that the samples from the minority modes have larger RLSs. Thus, when sampling from the RLS distribution, there is a higher probability of including the minority modes.

\noindent This paper is motivated by two situations where minority modes can occur: 
1) the observed empirical distribution is different from the true distribution (biased data), and the data needs to be rebalanced. 2) The observed empirical distribution approximates the true distribution sufficiently well, but minority modes consist out of infrequent but very important points, e.g. rare diseases in a medical dataset.\\

\noindent\textbf{Contribution.} 
When training classical GANs, an empirical expectation of a loss $\mathbb{E}_{x\sim p_d}[\mathcal{L}(x)]$ is optimized in the context of a min-max problem.  In this work, we propose a sampling procedure that promotes sampling out of minority modes by using ridge leverage scores.
The common algorithmic procedure simulates the empirical distribution over the dataset $\mathcal{D}=\{x_1, \dots, x_n\}$ by uniformly sampling over this set. We intentionally \emph{bias} or \emph{distort} this process by sampling $x_i$ with probability
$
p(x_i) \propto \ell_i,
$
where $\ell_i$ is the $i$-th ridge leverage score, defined in Section~\ref{sec:RLS_Sampling}.
Empirical evidence shows that our procedure rebalances the training distribution, as a result, the GAN model generates samples more uniformly over all modes. 
RLS sampling can easily be applied to any GAN. In particular, using our procedure in combination with a state-of-the-art method for complete mode coverage~\cite{zhong2019rethinking} shows a clear improvement. Finally, RLS sampling is combined with  BuresGAN~\cite{demeulemeester} and a state-of-the-art StyleGAN2 with differentiable data augmentations~\cite{zhao2020differentiable}, which in both cases improves mode coverage\footnote{Code and supplementary at \url{https://github.com/joachimschreurs/RLS_GAN}}.  \\

\noindent\textbf{Related work.} Several works discuss alternative sampling strategies in machine learning.
In the context of risk-averse learning, the authors of~\cite{curi2019adaptive} discuss an adaptive sampling algorithm that performs a stochastic optimization of the Conditional Value-at-Risk (CVaR) of a loss distribution. This strategy promotes models which do not only perform well on average but also on rare data points. 
In the context of generative models, AdaGAN~\cite{tolstikhin2017adagan} is a boosting approach to solve the missing mode problem, where at every step a new component is added into a mixture model by running the GAN training algorithm on a re-weighted sample. A supervised weighting strategy for GANs is proposed in~\cite{diesendruck2019importance}.
In this paper, we compare against two state-of-the-art GANs that combat mode collapse, PacGAN~\cite{lin2018pacgan} and BuresGAN~\cite{demeulemeester}. PacGAN uses a procedure called packing. This modifies the discriminator to make decisions based on multiple samples from the same class, either real or artificially generated. In BuresGAN, an additional diversity metric in the form of the Bures distance between real and fake covariance matrices is added to the generator loss. Note that these methods tackle the traditional mode collapse problem, i.e., the data does not include minority modes. 
In~\cite{zhang2017determinantal,zhang2019active}, it was shown that the convergence speed of stochastic gradient descent can be improved by actively selecting mini-batches using DPPs. In~\cite{pmlr-v119-sinha20b}, coreset-selection is used to create mini-batches with a ‘coverage’ similar to that of the large batch – in particular, the small batch tries to ‘cover’ all the same modes as are covered in the large batch. 

Before proceeding further, we discuss two main competitors more in-depth. The authors of~\cite{diesendruck2019importance} propose a solution to reduce selection bias in training data named Importance Weighted Generative Networks. A rescaling of the empirical data distribution is performed during training by employing a weighted Maximum Mean Discrepancy (MMD) loss such that the regions where the observed and the target distributions differ are penalized more. 
Each sample $i\in\{1,\dots,n\}$ is scaled by $1/M(x_i)$, where $M$ is the known or estimated Radon-Nykodym derivative between the target and observed distribution.
 A version of the vanilla GAN with importance weighting is introduced (IwGAN), as well as the weighting combined with MMDGAN (IwMmdGAN).  Another approach to complete mode coverage by~\cite{zhong2019rethinking} and dubbed MwuGAN in this paper, iteratively trains a mixture of generators. At each iteration, the sampling probability is pointwise normalized so that the probability to sample a missing mode is increased. Hence, this generates a sequence of generative models which constitutes the mixture. More precisely, a weight $w_i>0$ is given for each $i\in \{1,\dots,n\}$ and initialized such that $w_i = p(x_i)$ for some distribution\footnote{In~\cite{zhong2019rethinking}, this initial distribution is uniform. We discuss in Section~\ref{sec:synth} a choice of weights based on RLSs and initialize MwuGAN with the normalized RLSs in~\eqref{eq:RLS_prob}.} $p$. Next, a generative model is trained and the probability density $p_g(x_i)$ of each $i\in \{1,\dots,n\}$ is computed. If $p_g(x_i)<\delta p(x_i)$ for some threshold value $\delta\in (0,1)$, the weight is updated as follows: $w_i \leftarrow 2 w_i$, otherwise the weight is not updated. The probability is then recalculated as follows: $p(x_i) = w_i/\sum_j w_j$ for each $i\in \{1,\dots,n\}$. Another generative model is then trained by using $p(x_i)$ and the procedure is repeated. \\

\noindent\textbf{Classical approach.} 
A GAN consists of a discriminator $D: \mathbb{R}^d \to \mathbb{R}$ and a generator $G: \mathbb{R}^\ell \to \mathbb{R}^d$ which are typically defined by neural networks, and parametrized by real vectors. The value $D (x)$ gives the probability that $x$ comes from the empirical distribution, while the generator $G$ maps a point $z$ in the latent space $\mathbb{R}^\ell$ to a point in input space $\mathbb{R}^d$. A typical training scheme for a GAN consists in solving, in an alternating way, the following problems:
\begin{equation}
\label{eq:GAN}
\begin{aligned}
&\max _{D} \mathbb{E}_{x \sim p_{d}}[\log(D(x))]+\mathbb{E}_{\tilde{x} \sim p_{g}}[\log(1-D(\tilde{x}))], \\
&\min _{G} -\mathbb{E}_{\tilde{x} \sim p_{g}}[\log(D(\tilde{x}))],
\end{aligned}
\end{equation}
which include the vanilla GAN objective associated with the cross-entropy loss.  In~\eqref{eq:GAN}, the first expectation is over the empirical data distribution  $p_{d}$ and the second is over the generated data distribution $p_g$, implicitly given by the mapping by $G$ of the latent prior distribution $\mathcal{N}(0,\mathbb{I}_\ell)$.
%Let $\mathcal{D} = \{x_1,\dots, x_n\}$.
The data distribution $p_{d}$  is estimated using the empirical distribution over the training data
$\hat{p}_{d}(x) =\frac{1}{n} \sum_{x_{i} \in \mathcal{D}} \delta\left(x-x_{i}\right)$ as follows: $\mathbb{E}_{p_{d}(x)}[\mathcal{L}(x)] \approx \mathbb{E}_{\hat{p}_{d}(x)}[\mathcal{L}(x)]=\frac{1}{n} \sum_{x_{i} \in \mathcal{D}} \mathcal{L}\left(x_{i}\right)$, where $\mathcal{L}$ is a general loss function. As noted by~\cite{tripp2020sample}, positive weights $w_{i}$ for $1\leq i\leq n$ can be used to construct a weighted empirical distribution $\hat{p}_d^{w}(x)=\sum_{x_{i} \in \mathcal{D}} w_{i} \delta\left(x-x_{i}\right),$ then one can apply a weighting strategy to use samples distributed according to $\hat{p}(x)$ to estimate quantities with respect to $\hat{p}_d^{w}(x)$ as follows:
\begin{equation}
\mathbb{E}_{\hat{p}_d^{w}(x)}[\mathcal{L}(x)] = \mathbb{E}_{\hat{p}_d(x)}\left[\frac{\hat{p}_d^{w}(x)}{\hat{p}_d(x)} \mathcal{L}(x)\right] 
 =\sum_{x_{i} \in \mathcal{D}} w_{i} \mathcal{L}\left(x_{i}\right). \label{eq:weightedobj}
\end{equation}
A stochastic procedure is applied for minimizing the above expectation over $\hat{p}_d^{w}$. In this paper, mini-batches are sampled according to the distribution $\hat{p}_d^{w}$ with $w_i$ given by the normalized RLSs~\eqref{eq:RLS_prob} for $1\leq i\leq n$.

\section{Sampling with Ridge Leverage Scores \label{sec:RLS_Sampling}}

\begin{figure}[h]
	\centering
    \begin{minipage}{.25\textwidth}
        \centering
        \includegraphics[width=1\linewidth]{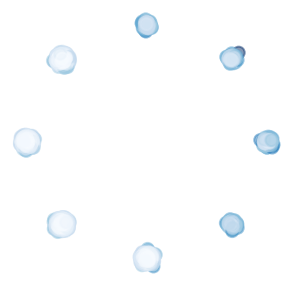}
    \end{minipage}%
    \qquad 
    \begin{minipage}{0.25\textwidth}
        \centering
        \includegraphics[width=1\linewidth]{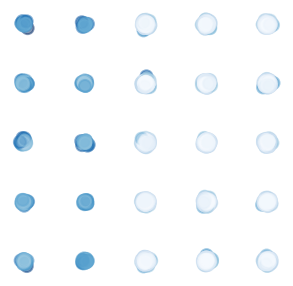}
    \end{minipage}
    \caption{RLS distribution using a Gaussian feature map with $\sigma = 0.15$ and regularization $\gamma = 10^{-3}$ on the unbalanced \textsc{Ring} (left) and \textsc{Grid} (right) data. The darker the shade, the higher the RLS. Dark modes correspond to minority modes. }
	\label{fig:RLS_Synth}
\end{figure}

We propose to use a sampling procedure based on ridge leverage scores (RLSs)~\cite{alaoui2015fast,pmlr-v32-ma14}. RLSs correspond to the correlation between the singular vectors of a matrix and the canonical basis elements. The higher the score, the more \emph{unique} the point. A sample from a minority mode would thus get a higher RLS. These RLSs are used to bias the sampling, which in turn results in a more uniform sampling over all the modes, regardless of the original weight of the mode in the data distribution. Given a feature map $\varphi(\cdot)$, the corresponding kernel function is $K(x,y) = \varphi(x)^\top \varphi(y)$.  Let the regularization parameter be $\gamma>0$. Then, the $\gamma$-RLSs are defined for all $1\leq i\leq n$ as:
\begin{equation}
\ell_{i}(\gamma) =\left(K(K+ n\gamma \mathbb{I})^{-1}\right)_{i i}= \varphi(x_i)^\top(C + n\gamma \mathbb{I})^{-1}\varphi(x_i), \label{eq:RLS_prob}
\end{equation}
where $C = \sum_{i=1}^n \varphi(x_i)\varphi(x_i)^\top$ and $K_{ij} = \varphi(x_i)^\top \varphi(x_j)$ for $1\leq i,j\leq n$. 
They have both a primal and a dual expression that can be leveraged when the size of the feature map or batch-size respectively are too large. When both the batch-size and feature map dimensions are large, one can rely on fast and reliable approximation algorithms with guarantees such as RRLS~\cite{musco2017recursive} and BLESS~\cite{rudi2018fast}. The role of $\gamma>0$ is to filter the small eigenvalues of $K$ in the spirit of Tikhonov regularization. RLSs induce the probability distribution: $p_i = \ell_i/\sum_{j=1}^n \ell_j,$ for $1\leq i\leq n$, which is classically used in randomized methods~\cite{alaoui2015fast}. 
Figure~\ref{fig:RLS_Synth} illustrates the interpretation of RLSs on two artificial datasets used in this paper. The datasets consist of a mixture of Gaussians. In the \textsc{Ring} example, the first $4$ modes are minority modes (starting on top and going further clockwise). In the \textsc{Grid} example, the first $10$ modes are minority modes (starting left). Similar to the first illustration (see Figure~\ref{fig:first example}), large RLSs are associated with minority modes. More information on the artificial datasets is given in Section~\ref{sec:synth}.

RLS sampling has a rich history in kernel methods and randomized linear algebra but has not been used in the context of GANs. One of the key contributions of this paper is to illustrate the use of RLSs in this setting. To do so, we propose the use of different feature maps so that RLS sampling can be used both for low dimensional and high dimensional data such as images.  In what follows, the feature map construction is first discussed. Next, two approximation schemes are introduced.\\

\noindent\textbf{Choice of the feature map.}
Three choices of feature maps are considered in this paper to compute leverage scores:
\begin{itemize}
    \item Fixed implicit feature map. In low dimensional examples, the feature map can be chosen implicitly such that it corresponds to the Gaussian kernel: $\varphi(x)^\top \varphi(y) = \exp \left(-\left\|x-y\right\|^{2}/\sigma^{2}\right)$, the bandwidth $\sigma$ is a hyperparameter. 
    \item Fixed explicit feature map. For image-based data, more advanced similarity metrics are necessary. Therefore, the next-to-last layer of a pre-trained classifier, e.g. the Inception network, is used to extract meaningful features. Note that the classifier does not need to be trained on the exact training dataset, but simply needs to extract useful features. 
     \item Discriminator-based explicit feature map. The feature map can be obtained from the next-to-last layer of the discriminator.  Let $D (x) = \sigma(w^\top \varphi_D(x))$, where $w\in \mathbb{R}^m$ contains the last dense layer's weights and $\sigma$ is the sigmoid function. This feature map is useful in situations where no prior knowledge is available about the dataset. 
\end{itemize}
For a fixed feature map, the RLSs only need to be calculated once before training. The discriminator-based explicit feature map changes throughout the training. Therefore the RLSs are recalculated at every step. Nonetheless, due to approximation schemes which are discussed hereafter, the computational cost stays low. The full algorithm is given in Supplementary Material.

\subsection{Approximation schemes}

Current day models are high dimensional, e.g., a DCGAN yields a feature space $\mathbb{R}^m$ of high dimension $m = 10^3$. Moreover, the size of datasets is commonly thousands up to millions of images.
Therefore, two approximation schemes are proposed to speed up the computation of RLSs when using explicit feature maps: 
\begin{itemize}
    \item For the discriminator-based explicit feature map, we propose a two-stage sampling procedure in combination with a Gaussian sketch to reduce the dimension of high dimensional feature maps. 
    \item For the fixed explicit feature map, the well-known UMAP is used to reduce the dimensionality~\cite{mcinnes2018umap}.
\end{itemize}

\noindent\textbf{Two-stage sampling procedure.} For the explicit discriminator-based feature map, $\varphi_D$ has to be re-calculated at each training step. To speed up the sampling procedure, we propose a sampling procedure in two stages. First, a subset of the data is uniformly sampled, e.g. equal to $20$ times the desired batch size. Afterward, the RLSs are calculated only for the uniformly sampled subset, which are then used to sample the final batch used for training. This two-stage sampling procedure is similar to the core-set selection used in smallGAN~\cite{pmlr-v119-sinha20b}. A first difference is that core-sets are selected by combining a Gaussian sketch and a greedy selection in~\cite{pmlr-v119-sinha20b}, while we use a randomized approach. Second, in this reference, cores-sets are used to reduce the batch size to improve scalability. In contrast, RLS sampling is used here to bias the empirical distribution. \\

\noindent\textbf{Sketching the discriminator feature map.}  Gaussian sketching is a commonly used method to reduce data dimension and was also used in~\cite{pmlr-v119-sinha20b} and~\cite{yang2015deep} to reduce the dimension of large neural nets. Let $S$ be a sketching matrix of size $m\times k$ such that $S= A/\sqrt{k}$ with $A$ a matrix with i.i.d. zero-mean standard normal entries. Consider the following random projection: let a batch be $\{i_1,\dots, i_b\}\subset \{1,\dots, n\}$. A random projection of this batch in feature space is then defined as follows:
\begin{equation}
\varphi(x_{i_\ell}) = S^\top\varphi_D(x_{i_\ell}) \in \mathbb{R}^{k},\label{eq:Proj}
\end{equation}
for all $\ell\in\{1,\dots, b \}$. This random projection preserves approximately (squared) pairwise distances in the dataset and is motivated by an isometric embedding result in the spirit of Johnson-Lindenstrauss lemma. Let $0<\epsilon < 1$ and any integer $b>0$. Let $k$ be an integer such that $k\geq 4(\epsilon^2/2-\epsilon^3/3)^{-1} \log b$. Then, for any set $\{\mathsf{x}_1, \dots, \mathsf{x}_b\}$ in $\mathbb{R}^m$ there is a map $f:\mathbb{R}^m\to \mathbb{R}^k $ such that for any $\ell, \ell'\in \{1, \dots, b\}$ we have
$$
(1-\epsilon) \|\mathsf{x}_\ell -\mathsf{x}_{\ell'} \|^2_2 \leq \| f(\mathsf{x}_\ell) -f(\mathsf{x}_{\ell'}) \|^2_2 \leq (1+\epsilon) \|\mathsf{x}_\ell -\mathsf{x}_{\ell'}  \|^2_2.
$$
The idea of this work is to consider the set of points given by the batch in the discriminator feature space $\mathsf{x}_\ell = \varphi_D(x_{i_\ell})$ for $1\leq \ell\leq b$.
It is proved in~\cite{Dasgupta} that $f$ exists and can be obtained with high probability with a random projection of the form~\eqref{eq:Proj}. For a more detailed discussion, we refer to~\cite{IsometricSketching}. \\

\noindent\textbf{Dimensionality reduction of the fixed explicit feature map by UMAP.}  The Gaussian sketch is a simple and fast method to reduce the dimension of the feature map. This makes it a perfect candidate to reduce the dimension of the proposed discriminator feature map, which has to be recalculated at every iteration. Unfortunately, this speed comes at a price, namely, the gaussian sketch is deemed too simple to reduce the dimension of highly complex models like the Inception network. Therefore, UMAP is proposed~\cite{mcinnes2018umap}. This non-linear dimensionality reduction technique can extract more meaningful features. UMAP is considerably slower than the Gaussian sketch, therefore the use of UMAP is only advised for a fixed feature map like a pre-trained classifier or the Inception network, where the RLSs are only calculated once before training.

\section{Numerical experiments}

The training procedure is evaluated on several synthetic and real datasets, where we artificially introduce minority modes. The GANs are evaluated by analyzing the distribution of the generated samples. Ideally, the models should generate samples from every mode as uniformly as possible. A re-balancing effect should be visible. The proposed methods are compared with vanilla GAN, PacGAN, MwuGAN, BuresGAN, IwGAN, and IwMmdGAN. In particular,  MwuGAN outperforms AdAGAN~\cite{tolstikhin2017adagan} on complete mode coverage problems~\cite{zhong2019rethinking}. 
BuresGAN~\cite{demeulemeester} promotes a matching between the covariance matrices of real and generated data in a feature space defined thanks to the discriminator.
Recall that the discriminator is $D (x) = \sigma(w^\top \varphi_D(x))$, where $w$ is a weight vector and the sigmoid function is denoted by $\sigma$. The normalization $\bar{\varphi}_D(x) = \varphi_D(x)/\|\varphi_D(x)\|_2$ is used, after the centering of $\varphi_D(x)$. Then, the covariance matrix is defined as follows:  $C(p) =  \mathbb{E}_{x\sim p}[\bar{\varphi}_D(x)\bar{\varphi}_D(x)^\top]$.
The real data and generated data covariance matrices are denoted by $C_d =C(p_d)$ and $C_g =C(p_g)$, respectively. In BuresGAN, the Bures distance is added to the generator loss: $\min_{G} -\mathbb{E}_{\tilde{x} \sim p_{g}}[\log(D(\tilde{x}))] + \lambda \mathcal{B}(C_{d},C_{g})^2$,
with the Bures distance
$\mathcal{B}\left({C_{d}}, {C_{g}}\right)^{2} =  \mathrm{Tr}({C_{d}}+{C_{g}}-2\left(C_{d}C_{g}\right)^{\frac{1}{2}
}),
$
depends implicitly on $\varphi_D(x)$ (see e.g.,~\cite{Massart}). The loss of the discriminator remains the same. \\

\noindent\textbf{Overview of proposed methods.}
The RLS sampling procedure can easily be integrated into any GAN architecture or model. In this spirit, we used RLS sampling with a classical vanilla GAN (\textbf{RLS GAN}) and BuresGAN (\textbf{RLS BuresGAN}). The classical BuresGAN has been shown to outperform competitors in mode collapse problems~\cite{demeulemeester}. We noticed empirically that RLS BuresGAN outperformed RLS GAN on the synthetic data (the comparison is shown in Supplementary Material). Therefore we only continue with RLS BuresGAN in the rest of the experiments. Likewise, RLS sampling is combined with MwuGAN, which is considered state-of-the-art in complete mode coverage. The method, called \textbf{RLS MwuGAN}, uses RLSs as initial starting weights to sample as opposed to uniform weights. Besides the initialization, the method remains unchanged. The number of generators in the mixture is always displayed in brackets. Unless specified otherwise, the models are trained for $30$k iterations with a batch size of $64$, by using the Adam~\cite{kingma2014adam} optimizer. Unless specified otherwise, we report the means and standard deviations for 10 runs. The largest mean is depicted in black, a $^\star$ represents significance using a one-tailed Welch's t-test between the best performing proposed model and best performing competitor at a $0.05$ confidence level. Further information about the used architectures, hyperparameters, and timings are given in Supplementary Material. 

\begin{figure*}[h]
	\centering
    \begin{minipage}{.24\textwidth}
        \centering
        \includegraphics[width=1\linewidth]{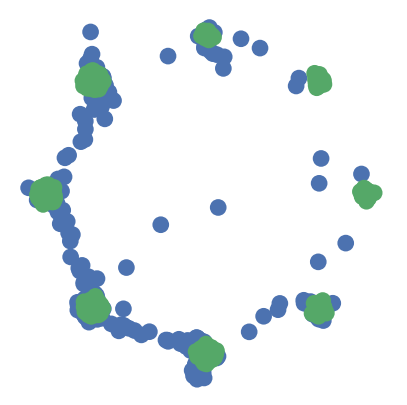}
    \end{minipage}%
    \begin{minipage}{0.24\textwidth}
        \centering
        \includegraphics[width=1\linewidth]{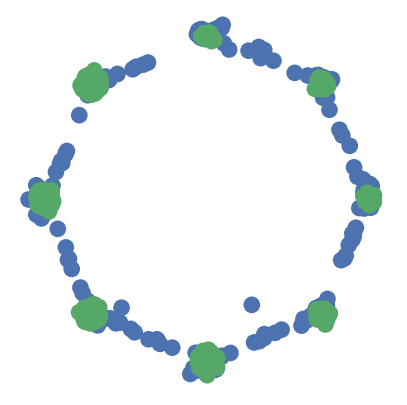}
    \end{minipage}
   \begin{minipage}{.24\textwidth}
        \centering
        \includegraphics[width=1\linewidth]{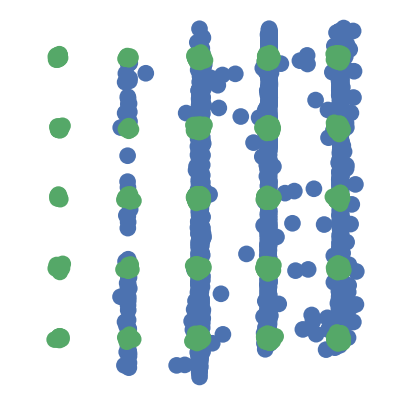}
    \end{minipage}%
    \begin{minipage}{0.24\textwidth}
        \centering
        \includegraphics[width=1\linewidth]{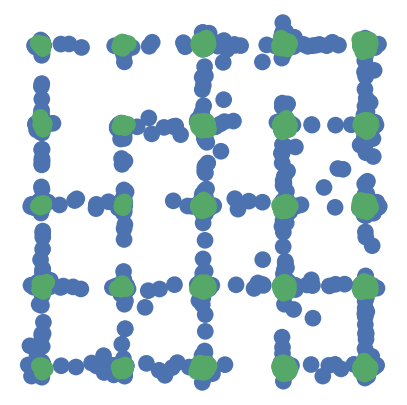}
    \end{minipage}
    \caption{Visualization of the generation quality on \textsc{Ring} and \textsc{Grid}. Each column shows $2.5$k samples from the trained generator in blue and $2.5$k samples from the true distribution in green. The vanilla GAN (first and third) does not cover the minority modes. This is not the case for the RLS BuresGAN Discr. (second and fourth).}
	\label{fig:Samples_Synth}
\end{figure*}

\subsection{Synthetic data \label{sec:synth}}
Unbalanced versions of two classical synthetic datasets are generated: an unbalanced ring with $4$ minority modes (\textsc{Ring}) and an unbalanced grid (\textsc{Grid}) with $10$ minority modes (see Figure~\ref{fig:RLS_Synth}).
\textsc{Ring} is a mixture of eight two-dimensional isotropic Gaussians in the plane with means $2.5\times(\cos((2\pi/8)i), \sin((2\pi/8)i))$ and std $0.05$ for $i \in \{1,\dots,8\}$. The probability of sampling from the first $4$ consecutive Gaussians is only $0.05$ times the probability of sampling from the last $4$ modes.
\textsc{Grid} is a mixture of $25$ two-dimensional isotropic normals with standard deviation $0.05$ and with means on a square grid with spacing $2$. The first rectangular blocks of $2\times 5$ adjacent modes are depleted with a factor $0.05$.

\begin{table}[h]
    \caption{Experiments on the synthetic datasets \textsc{Ring} and \textsc{Grid}. 
    Two RLS BuresGAN are considered: RLSs calculated with the Gaussian kernel (Gauss.) and the next-to-last layer of the discriminator (Discr.).
     RLS MwuGAN is initialized with RLSs using an implicit feature map associated with the Gaussian kernel.}
    \label{tab:Synthetic}
\begin{center}
 \resizebox{0.98\textwidth}{!}{
    \begin{tabular}{lc c c c }
        \toprule
          & \multicolumn{2}{c}{Ring with 8 modes } & \multicolumn{2}{c}{Grid with 25 modes}\\
          \cmidrule(lr){2-3}\cmidrule(lr){4-5}
          & Nb modes ($\uparrow$) & $\%$ in $3\sigma$ ($\uparrow$) & Nb modes ($\uparrow$) & $\%$ in $3\sigma$ ($\uparrow$)  \\
           \cmidrule(lr){2-2}  \cmidrule(lr){3-3}\cmidrule(lr){4-4} \cmidrule(lr){5-5}
         GAN  & $5.0(1.1)$  & $\bm{0.92}(0.02)^\star$  & $8.3(3.4)$  & $0.29(0.3)$    \\
         PacGAN2   & $5.4(1.4)$  & $\bm{0.92}(0.03)$   & $10.3(2.6)$   & $0.13(0.02)$   \\
         BuresGAN  & $5.8(1.4)$ & $0.76(0.27)$  & $16.7(0.9)$   & $\bm{0.82}(0.01)^\star$   \\ \midrule
         RLS GAN Gauss. & $7.4(0.9)$  & $0.86(0.05)$  & $13.8(7.4)$  & $0.51(0.32)$   \\ 
         RLS GAN Discr. & $7.6(0.8)$  & $0.90(0.02)$  &  $20.4(2.6)$  & $0.81(0.03)$   \\ 
         \midrule
         IwGAN & $\bm{8}(0)$ & $0.85(0.08)$ & $13.4(5.7)$ & $0.29(0.25)$ \\
         IwMmdGAN & $\bm{8}(0)$ & $0.84(0.02)$  & $1.7(5.1)$   & $0.03(0.05)$    \\ \midrule
         MwuGAN (15) & $7.9(0.3)$  &  $0.86(0.02)$ & $15.2(1.7)$  & $0.47(0.11)$      \\ 
         RLS MwuGAN (15) (ours) & $\bm{8}(0)$  & $0.84(0.06)$ & $22.3(1.9)$  & $0.60(0.1)$      \\\midrule
         RLS BuresGAN Gauss. (ours) & $\bm{8}(0)$  & $0.90(0.02)$  & $24.0(1.5)$  & $0.76(0.11)$   \\ 
         RLS BuresGAN Discr.  (ours) & $\bm{8}(0)$  & $0.90(0.02)$  & $\bm{24.4}(0.92)^\star$  & $0.78(0.06)$   \\ 
         \bottomrule
    \end{tabular}
    }
\end{center}
\end{table}

\noindent\textbf{Evaluation.} The evaluation is done by sampling $10$k points from the generator network. High-quality samples are within $3$ standard deviations of the nearest mode. A mode is covered if there are at least $50$ generated samples within $3$ standard deviations of the center of the mode. The knowledge of the full Radon-Nikodym derivative $M$ is given to IwMmdGAN and IwGAN by dividing the true probability of each sample (a Gaussian mixture with equal weights) by the adapted Gaussian mixture containing several minority modes. The results of the experiments are given in Table~\ref{tab:Synthetic}. For RLS sampling, we use a Gaussian kernel with bandwidth $\sigma = 0.15$, and the discriminator network as feature extractor, both with regularization parameter $\gamma = 10^{-3}$. The models are trained using a fully connected architecture (see Supplementary Material). As the models are rather simple, no dimensionality reduction is needed. 

\begin{figure}[h]
	\centering
	 \begin{minipage}{.32\textwidth}
        \centering
        \includegraphics[width=1\linewidth]{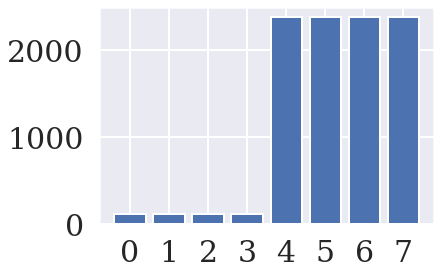}
    \end{minipage}%
    \begin{minipage}{0.32\textwidth}
        \centering
        \includegraphics[width=1\linewidth]{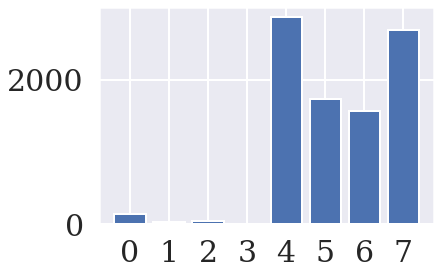}
    \end{minipage}
    \begin{minipage}{.32\textwidth}
        \centering
        \includegraphics[width=1\linewidth]{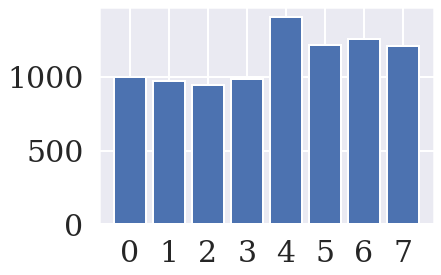}
    \end{minipage}%
    \caption{\textsc{Ring}. Number of training samples in each mode for the \textsc{Ring} dataset (left) Generated samples in each mode by a vanilla GAN (middle). Generated samples by RLS BuresGAN Discr. (right). A rebalancing effect is visible.}
	\label{fig:Samp_Ring}
\end{figure}

\noindent Generated samples from models trained with and without RLS sampling are displayed in Figure~\ref{fig:Samples_Synth}. One can clearly see that training a GAN with uniform sampling results in missing the first 4 minority modes. This is solved by using RLS sampling and can be interpreted by comparing the two sampling distributions on Figure~\ref{fig:Samp_Ring} (uniform) and  Figure~\ref{fig:RLS_Synth} (RLS). The RLSs are larger for samples in minority modes, which results in a more uniform mini-batch over all modes. Note that the RLS sampling procedure, given the feature map, is completely unsupervised and has no knowledge of the desired unbiased distribution. The evaluation metrics in Table~\ref{tab:Synthetic} confirm our suspicions.  Only methods designed for complete mode coverage can recover (almost) all modes for the \textsc{Ring} dataset. For the unbalanced \textsc{Grid}, only the proposed method has an acceptable performance. Our method even outperforms multiple generator architectures like MwuGAN and RLS MwuGAN, which are considerably more costly to train. Moreover, IwGAN, with full knowledge of $M$, is not capable of consistently capturing all modes. This was pointed out by the authors in~\cite{diesendruck2019importance}: the method may still experience high variance if it rarely sees data points from a class it wants to boost. 
\subsection{Unbalanced MNIST}
For this experiment, we create two unbalanced datasets out of MNIST.
The first modified dataset, named \textsc{unbalanced $012$-MNIST}, consists of only the digits $0,1$ and $2$.  The class $2$ is depleted so that the probability of sampling $2$ is only $0.05$ times the probability of sampling from the digit $0$ or $1$. The second dataset, named \textsc{unbalanced MNIST}, consists of all digits. The classes $0$,$1$,$2$,$3$, and $4$ are all depleted so that the probability of sampling out of the minority classes is only $0.05$ times the probability of sampling from the majority digits.  For these experiments, we use a DCGAN architecture. The following metrics are used for performance evaluation: the number of generated digits in each mode, which measures mode coverage, and the KL divergence~\cite{metz2016unrolled} between the classified labels of the generated samples and a balanced label distribution, which measures sample quality. The mode of each generated image is identified by using a MNIST classifier which is trained up to $98.43\%$ accuracy (see Supplementary Material). The metrics are calculated based on $10$k generated images for all the models.  
For the RLS computation, we use both the discriminator with a Gaussian sketch and the next-to-last layer of the pre-trained classifier with a UMAP dimensionality reduction as a feature map. For the \textsc{unbalanced $012$-MNIST}, both feature maps are reduced to $k = 25$ and the regularization parameter is $\gamma = 10^{-4}$. In the \textsc{unbalanced MNIST}, we take $k=10$ and $\gamma = 10^{-4}$.  An ablation study over different $k$ and $\gamma$ is given in Supplementary Material. We also compare the performance of the classical MwuGAN, initialized with uniform weights, with RLS MwuGAN where the weights are initialized by the RLSs calculated using the fixed explicit feature map with the same parameters mentioned above. Both methods contain a mixture of $15$ GANs, the experiments are repeated $3$ times for MwuGAN variants. 
\begin{figure}[h]
	\centering
    \begin{minipage}{0.24\textwidth}
        \centering
        \includegraphics[width=1\linewidth]{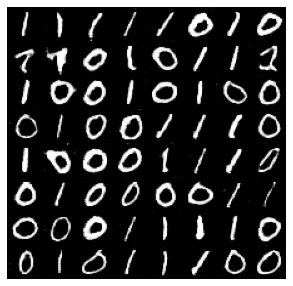}
    \end{minipage}
    \begin{minipage}{.24\textwidth}
        \centering
        \includegraphics[width=1\linewidth]{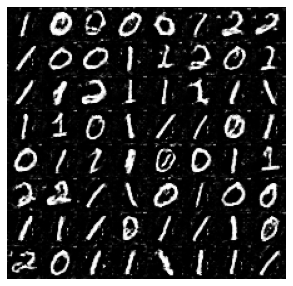}
    \end{minipage}
    \begin{minipage}{.24\textwidth}
        \centering
        \includegraphics[width=1\linewidth]{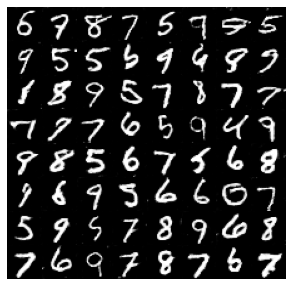}
    \end{minipage}
    \begin{minipage}{.24\textwidth}
        \centering
        \includegraphics[width=1\linewidth]{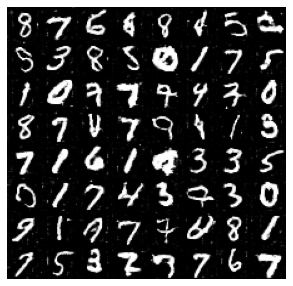}
    \end{minipage}%
    \caption{Generated images from \textsc{unbalanced 012-MNIST} by a vanilla GAN (first), by RLS BuresGAN Discr. (second) and generated images from \textsc{unbalanced MNIST} by a vanilla GAN (third), by RLS BuresGAN Class. (fourth). The minority digits are generated more frequently in the proposed methods that include RLS sampling.}
	\label{fig:Samp_MNIST_images}
\end{figure}
In our simulations, IwMmdGAN could not be trained successfully with a DCGAN architecture. The Radon-Nikodym derivative $M$, which is used by IwMmdGAN and IwGAN, is defined as follows: $M_i = 1$ for digits $0$ and $1$ and $M_i = 0.05$ for digits $2$, analogous for the \textsc{unbalancedMNIST} dataset. Only the proposed models trained with RLS sampling are capable of covering all modes consistently. The diversity of images generated by RLS BuresGAN can be visualized in Figure~\ref{fig:Samp_MNIST_images} where digits from minority modes appear more frequently.
A quantitative analysis of mode coverage and sample quality is reported in Tables~\ref{tab:012MNIST_samp} and \ref{tab:FullMNIST_samp}. In the \textsc{unbalanced $012$-MNIST} dataset, there is a clear advantage in using RLS sampling with BuresGAN since mode coverage and the KL divergence are improved compared to the other methods. The second best method is RLS MwuGAN which outperforms RLS with uniform starting weights in KL. For the more difficult \textsc{unbalanced MNIST} dataset, using the fixed explicit feature map to calculate the RLSs clearly outperforms other methods. 
\begin{table}[h]
    \caption{Experiments on the \textsc{unbalanced $012$-MNIST} dataset.  Two RLS BuresGAN are considered: RLSs calculated with an explicit feature map obtained from a pre-trained classifier (Class.) and the next-to-last layer of the discriminator (Discr.). RLS MwuGAN is initialized with RLSs using the explicit feature maps obtained from a pre-trained classifier. Minority modes are highlighted in black in the first row.}
    \label{tab:012MNIST_samp}
\begin{center}
    \begin{tabular}{l c c c c}
        \toprule
          & Mode 1 & Mode 2 & \textbf{Mode 3} & KL \\
          \cmidrule(lr){2-2}  \cmidrule(lr){3-3} \cmidrule(lr){4-4} \cmidrule(lr){5-5}
         GAN  & $4381(172)$  & $5412(179)$ &   $129(36)$ & $ 0.31(0.01)$  \\
         PacGAN2   & $4492(237)$  &  $5328(242)$ &  $123(29)$  & $0.32(0.01)$       \\
         BuresGAN & $4586(287)$   &  $5190(292)$ &  $142(19)$ & $0.30(0.01)$    \\ \midrule
         IwGAN &  $4368(295)$ & $5414(287)$  & $147 (32)$ & $ 0.32(0.01)$   \\ 
         IwMmdGAN & $34(12)$ & $0(0)$ & $69(12)$  & $0.56(0.10)$  \\ \midrule
         MwuGAN (15) & $4886(473)$  & $4865(466)$ & $176(14)$  &  $0.31(0.01)$       \\ 
         RLS MwuGAN (15) (ours) & $3982(218)$ & $4666(164)$ & $870(65)$  & $0.14(0.01)$  \\ \midrule
         RLS BuresGAN Class. (ours) & $3414(161)$ & $4862(134)$ & $1461(183)$ & $\bm{0.08}(0.01)^\star$ \\ 
         RLS BuresGAN Discr. (ours) & $5748(172)$ & $2416(268)$ & $\bm{1566}(293)^\star$ & $0.16(0.02)$  \\ 
         \bottomrule
    \end{tabular}
\end{center}
\end{table}
\begin{table}[h]
    \caption{Experiments on the \textsc{unbalanced MNIST} dataset. Two RLS BuresGAN variants are considered: RLSs calculated with an explicit feature map obtained from a pre-trained classifier (Class.) and the next-to-last layer of the discriminator (Discr.). RLS MwuGAN is initialized with RLSs using the explicit feature maps obtained from a pre-trained classifier. Only the number of samples in the minority modes are visualized. The number of samples in the remaining modes are given in Supplementary Material.}
    \label{tab:FullMNIST_samp}
\begin{center}
\resizebox{0.98\textwidth}{!}{
    \begin{tabular}{l c c c c c c c }
        \toprule
          & \textbf{Mode 1} & \textbf{Mode 2} & \textbf{Mode 3}  & \textbf{Mode 4}  & \textbf{Mode 5}  & KL \\
   \cmidrule(lr){2-2} \cmidrule(lr){3-3}\cmidrule(lr){4-4}\cmidrule(lr){5-5} \cmidrule(lr){6-6} \cmidrule(lr){7-7} 
         GAN  & $123(22)$ & $137(103)$ & $81(31)$ & $161(97)$ & $161(23)$ & $0.48(0.02)$  \\
         PacGAN2   & $109(29)$ & $142(70)$ & $89(23)$ & $147(100)$ & $152(40)$  & $0.48(0.02)$      \\
         BuresGAN &   $126(31)$ & $157(97)$ & $108(30)$ & $153(62)$ & $147(26)$  & $0.46(0.02)$  \\ \midrule
         IwGAN &  $117(33)$ & $139(33)$ & $97(31)$ & $212(69)$ & $154(34)$ & $0.46(0.02)$   \\ 
         IwMmdGAN &  $1(1)$ & $0(0)$ & $23(16)$ & $\bm{1140}(367)^\star$ & $3(3)$  & $1.92(0.1)$ \\\midrule
         MwuGAN (15) &  $144(29)$ & $113(28)$ & $146(13)$ & $172(18)$ & $167(28)$ & $0.46(0.02)$      \\ 
         RLS MwuGAN (15) (ours) & $336(47)$ & $283(32)$ & $191(23)$ & $381(38)$ & $276(33)$ & $0.30(0.02)$    \\ \midrule
         RLS BuresGAN Class. (ours) & $\bm{875}(112)^\star$ & $\bm{663}(122)^\star$ & $\bm{360}(198)^\star$ & $831(59)$ & $\bm{615}(82)^\star$  & $\bm{0.09}(0.01)^\star$ \\ 
         RLS BuresGAN Discr. (ours) & $235(62)$ & $183(141)$ & $264(44)$ & $255(109)$ & $219(54)$ & $0.37(0.02)$ \\ 
         \bottomrule
    \end{tabular}
    }
\end{center}
\end{table}
\subsection{Unbalanced CIFAR10}
We conclude this section with an experiment on colored images, namely the \textsc{CIFAR10} dataset. This highly diverse dataset contains $32 \times 32$ color images from $10$ different classes. We consider two unbalanced variations. The first modified dataset, named \textsc{unbalanced $06$-CIFAR10}, consists of only the classes $0$ and $6$ or images of airplanes and frogs respectively. The class $0$ is depleted with a factor $0.05$. The second dataset, named \textsc{unbalanced $016$-CIFAR10}, consists of the classes $0$,$1$ and $6$. Compared to the previous dataset, we add images from the class automobile. Now, the class $6$ consisting of frogs is depleted with a factor $0.05$. We show the improvement of RLS sampling in a StyleGAN2 with differentiable data augmentation (StyleGAN2 + Aug.)~\cite{zhao2020differentiable} \footnote{Code taken from \url{https://github.com/mit-han-lab/data-efficient-gans}}.  By clever use of various types of differentiable augmentations on both real and fake samples, the GAN can match the top performance on \textsc{CIFAR10} with only 20\% training data and is considered state-of-the-art. The StyleGAN2 models are trained for $156$k iterations with a mini-batch size of $32$ using `color, translation, and cutout' augmentations, which is suggested by the authors when only part of the \textsc{CIFAR10} dataset is used. All the other parameters remained the same, only the sampling strategy is changed to RLS sampling in RLS StyleGAN2 + Aug. For the RLS computation, we use the discriminator feature map with Gaussian sketching and a fixed explicit feature map given by the next-to-last layer of the Inception network where UMAP is used to reduce the dimension. For both the RLSs, the dimension is reduced to $k=25$ and the regularization parameter is $\gamma = 10^{-4}$.
The performance is assessed using 10k generated samples at the end of training by the Inception Score (IS) and the Fr\'echet inception distance (FID) between the generated fake dataset and the balanced dataset.
\begin{figure}[h]
	\centering
    \begin{minipage}{0.48\textwidth}
        \centering
        \includegraphics[width=1\linewidth]{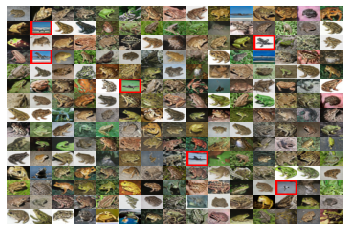}
    \end{minipage}
    \begin{minipage}{.48\textwidth}
        \centering
        \includegraphics[width=1\linewidth]{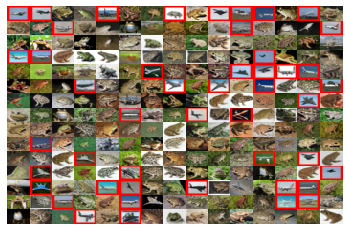}
    \end{minipage}%
    \caption{Generated images from \textsc{06-CIFAR10} by a  StyleGAN2 + Aug. (left) and by RLS  StyleGAN2 + Aug. (right). Including RLS sampling promotes sampling from the minority class. Generated samples classified as planes are marked by a red border.}
	\label{fig:Samp_CIFAR10_images}
\end{figure}
Mode coverage is evaluated by the number of generated samples in each class.
The class of a generated sample is evaluated by a trained \textsc{CIFAR10} classifier using a resnet56 type architecture~\cite{he2016deep} which is trained up to $93.77$\% accuracy\footnote{Classifier is available at \url{https://github.com/gahaalt/ResNets-in-tensorflow2}}. The results of the experiments are given in Table~\ref{tab:CIFAR10}, examples of generated images are given in Figure~\ref{fig:Samp_CIFAR10_images}. Including RLS sampling clearly improves the performance in unbalanced datasets, this is especially the case for the fixed feature map given by the Inception network. The minority mode is oversampled by approximately a factor $10$ or even $100$ in the case of the unbalanced \textsc{06-CIFAR10} and \textsc{016-CIFAR10} datasets respectively. Both the IS and FID also improve significantly. Note that the maximum achievable performance for IS and FID is lower when only a subset of classes is included, as pointed out by~\cite{sajjadi2018assessing}. %Nonetheless, the relative comparison remains instructive. 
\begin{table}[h]
    \caption{Experiments on the \textsc{unbalanced 06-CIFAR10} and \textsc{unbalanced 016-CIFAR10} dataset. Including RLS sampling in the StyleGAN2 + Aug. clearly improves the performance. Minority modes are highlighted in black in the second row.}
    \label{tab:CIFAR10}
\begin{center}
 \resizebox{0.98\textwidth}{!}{
    \begin{tabular}{l c c c c c c c c c}
        \toprule
        & \multicolumn{4}{c}{06-CIFAR10} & \multicolumn{5}{c}{016-CIFAR10} \\
        \cmidrule(lr){2-5}  \cmidrule(lr){6-10} 
         & \textbf{Mode 1} & Mode 2 & IS ($\uparrow$) & FID ($\downarrow$) & Mode 1 & Mode 2 & \textbf{Mode 3} & IS ($\uparrow$) & FID ($\downarrow$)\\ 
\cmidrule(lr){2-2} \cmidrule(lr){3-3}\cmidrule(lr){4-4}\cmidrule(lr){5-5} \cmidrule(lr){6-6} \cmidrule(lr){7-7} \cmidrule(lr){8-8} \cmidrule(lr){9-9} \cmidrule(lr){10-10} 
         StyleGAN2 + Aug.  & $261$  &  $9500$  & $4.8$ & $67.5$ & $4526$ &  $5206$ & $18$ & $4.3$ & $48.8$  \\
         RLS StyleGAN2 + Aug. Disc. (ours) & $994$ & $8659$ & $5.7$ & $46.4$ & $4449$ & $5132$ & $139$ & $4.6$ & $44.4$\\
         RLS StyleGAN2 + Aug. Class. (ours) & $\bm{2438}$ & $7212$ & $\bm{6.2}$ & $\bm{31.3}$ & $4156$ & $4393$ & $\bm{1155}$ & $\bm{5.7}$ & $\bm{27.2}$  \\
         \bottomrule
    \end{tabular}
    }
\end{center}
% \vspace{-10ex}
\end{table}

\section{Conclusion}
We introduced the use of RLS sampling for training GANs. This `diverse' sampling procedure was motivated by a notion of complete mode coverage in the presence of minority modes. RLS sampling is easy to integrate into any GAN model. Three feature maps have been discussed. An implicit feature map performs well for low-dimensional data. A fixed explicit feature map, such as a pre-trained classifier, achieves good results in high-dimensional cases. Lastly, the discriminator can be used as a feature map when no prior knowledge exists about the data. Two approximation methods for the explicit feature maps are also discussed: dimensionality reduction of explicit feature maps and a two-stage sampling procedure to efficiently speed up online RLS computation. We demonstrated empirically that the use of RLS sampling in GANs successfully combats the missing mode problem.  \\

\noindent \textbf{Acknowledgments}.
\footnotesize{EU: ERC Advanced Grants(787960, 885682). This paper reflects only the authors' views and the Union is not liable for any use that may be made of the contained information. Research Council KUL: projects C14/18/068,  C16/15/059, C3/19/053, C24/18/022, C3/20/117, Industrial Research Fund: 13-0260, IOF/16/004; Flemish Government: FWO: projects: GOA4917N, EOS Project no G0F6718N (SeLMA), SBO project S005319N, Infrastructure project I013218N, TBM Project T001919N; PhD Grants (SB/1SA1319N, SB/1S93918, SB/1S1319N), EWI: the Flanders AI Research Program. VLAIO: Baekeland PhD (HBC.20192204) and Innovation mandate (HBC.2019.2209), CoT project 2018.018. Foundation ‘Kom op tegen Kanker’, CM (Christelijke Mutualiteit). Ford KU Leuven Research Alliance Project KUL0076.}

%%%%%%%%%%%%%%%%%%%%%%%%%%%%%%%%%%

%\FloatBarrier
\bibliography{References}
\bibliographystyle{unsrt}

%%%%%%%%%%%%%%%%%%%%%%%%%%%%%%%%
\appendix
\section{Organization}
In Section~\ref{FullTable}, the remaining results from the \textsc{unbalanced MNIST} experiment are given. In Section~\ref{A}, the effect of the ridge parameter $\gamma$ and dimensionality reduction size $k$ is discussed. Then, in Section~\ref{B}, the training times of our simulations are reported. Section~\ref{C} provides further information about the distribution of generated samples in the artificial datasets. Finally, Section~\ref{D} describes the experimental setting, the proposed algorithm, and  architectures used in the paper.

\section{Extra table \textsc{unbalanced MNIST} dataset\label{FullTable}}

For the \textsc{unbalanced MNIST} experiment, only the number of samples in the minority modes are given in the main part. The number of samples for the remaining modes is given in Table~\ref{tab:FullMNIST_supp}. 

\begin{table}[h]
    \caption{Experiments on the \textsc{unbalanced MNIST} dataset. Two variants of RLS BuresGAN are trained. First, RLSs are calculated with an explicit feature map obtained from a pre-trained classifier (Class.). The second version uses the next-to-last layer of the discriminator (Discr.) to compute the RLSs. The RLS MwuGAN is initialized using the RLSs with the explicit feature maps obtained from a pre-trained classifier. }
    \label{tab:FullMNIST_supp}
\begin{center}
\resizebox{0.98\textwidth}{!}{
    \begin{tabular}{l c c c c c c c }
        \toprule
            & Mode 6 & Mode 7 & Mode 8 & Mode 9 & Mode 10  \\
   \cmidrule(lr){2-2} \cmidrule(lr){3-3}\cmidrule(lr){4-4}\cmidrule(lr){5-5} \cmidrule(lr){6-6}
         GAN   & $1831(135)$ & $1835(32)$ & $2009(69)$ & $1747(46)$ & $1915(143)$   \\
         PacGAN2  & $1767(116)$ & $1864(16)$ & $1983(105)$ & $1847(25)$ & $1900(110)$    \\
         BuresGAN  & $1762(96)$ & $1870(38)$ & $1979(110)$ & $1782(29)$ & $1916(68)$  \\ \midrule
         IwGAN & $1833(74)$ & $1837(97)$ & $1902(79)$ & $1783(64)$ & $1928(74)$   \\ 
         IwMmdGAN & $67(45)$ & $0(0)$ & $4(3)$ & $8741(430)$ & $22(10)$  \\\midrule
         MwuGAN (15) &  $1638(137)$ & $1986(114)$ & $1807(96)$ & $2026(22)$ & $1801(123)$      \\ 
         RLS MwuGAN (15) (ours) &  $1616(17)$ & $1513(144)$ & $2018(78)$ & $1576(30)$ & $1811(34)$   \\ \midrule
         RLS BuresGAN Class. (ours) & $1311(178)$ & $1137(116)$ & $1743(79)$ & $1069(67)$ & $1396(100)$  \\ 
         RLS BuresGAN Discr. (ours)  & $2012(97)$ & $1839(53)$ & $1608(155)$ & $1841(55)$ & $1545(115)$  \\ 
         \bottomrule
    \end{tabular}
}
\end{center}
\end{table}

\FloatBarrier
\section{Effect of the regularization parameter in RLS \label{A}}

\subsection{Synthetic data}

We display in Table~\ref{tab:Synt_ablation_disc} an ablation study for different ridge regularization parameters $\gamma>0$ on the synthetic \textsc{Ring} and \textsc{Grid}. A lesson from Table~\ref{tab:Synt_ablation_disc} is that the performance on the synthetic datasets does not vary much if different values of $\gamma$ are chosen in the case of the discriminator feature map.  Notice that, for $\gamma = 10^{-3}$ and $\gamma = 10^{-4}$, there is a small improvement in sample quality for \textsc{Ring} and in terms of mode coverage for \textsc{Grid} with the Gaussian kernel feature map.
\begin{table}[h]
    \caption{Ablation study  over the parameter $\gamma$ on the synthetic datasets for RLS BuresGAN with a  discriminator feature map (Discr.) and Gaussian feature map (Gauss.) with bandwidth $\sigma = 0.15$. No significant difference can be seen from the reported performance as the regularization parameter varies.}
    \label{tab:Synt_ablation_disc}
\begin{center}

    \begin{tabular}{l c c c c c }
        \toprule
      \multirow{2}{*}{Feature map}    & \multirow{2}{*}{$\gamma$} & \multicolumn{2}{c}{Ring with 8 modes } & \multicolumn{2}{c}{Grid with 25 modes}\\
          \cmidrule(lr){3-4}\cmidrule(lr){5-6}
          & &  Nb modes & $\%$ in $3\sigma$ & Nb modes & $\%$ in $3\sigma$  \\
          \cmidrule(lr){1-1} \cmidrule(lr){2-2} \cmidrule(lr){3-3}\cmidrule(lr){4-4} \cmidrule(lr){5-5} \cmidrule(lr){6-6}
  \multirow{3}{*}{Discr.}    &   $10^{-2}$ & $8 (0)$ & $0.92(0.02)$ & $24.0(1.3)$ & $0.79(0.03)$  \\
       &   $10^{-3}$ & $8 (0)$ & $0.90(0.02)$ & $24.4(0.9)$ & $0.78(0.06)$  \\
       &   $10^{-4}$ & $8 (0)$ & $0.90(0.01)$ & $24.5(0.8)$ & $0.77(0.07)$   \\ \midrule
\multirow{3}{*}{Gauss.} &  $10^{-2}$ & $8 (0)$ & $0.83(0.10)$ & $19.9(1.9)$ & $0.76(0.09)$  \\
         & $10^{-3}$ & $8 (0)$ & $0.90(0.02)$ & $24.0(1.5)$ & $0.76(0.11)$  \\
         & $10^{-4}$ & $8 (0)$ & $0.87(0.04)$ & $24.8(0.6)$ & $0.78 (0.03)$   \\        
         \bottomrule
    \end{tabular}

\end{center}
\end{table}

\begin{figure}[h]
	\centering
    \begin{minipage}{.24\textwidth}
        \centering
        \includegraphics[width=1\linewidth]{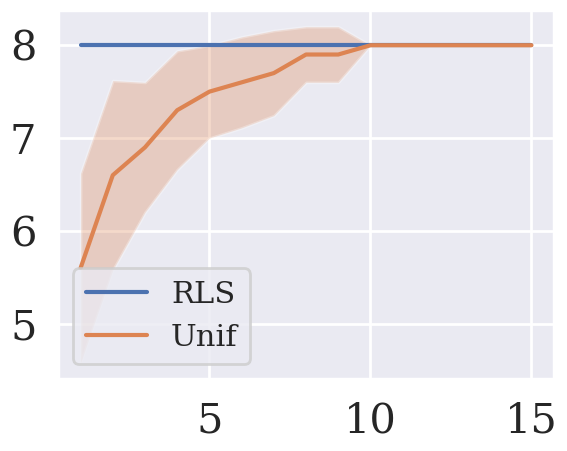}
    \end{minipage}%
    \begin{minipage}{0.24\textwidth}
        \centering
        \includegraphics[width=1\linewidth]{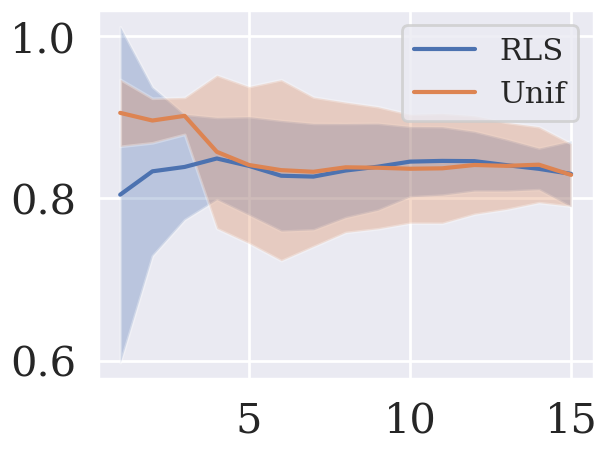}
    \end{minipage}
    \begin{minipage}{.24\textwidth}
        \centering
        \includegraphics[width=1\linewidth]{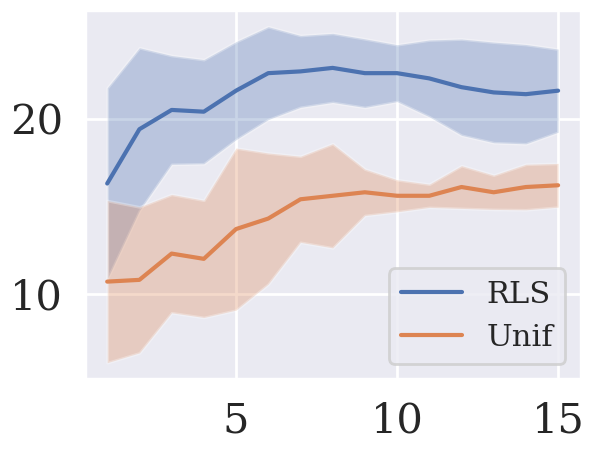}
    \end{minipage}%
    \begin{minipage}{.24\textwidth}
        \centering
        \includegraphics[width=1\linewidth]{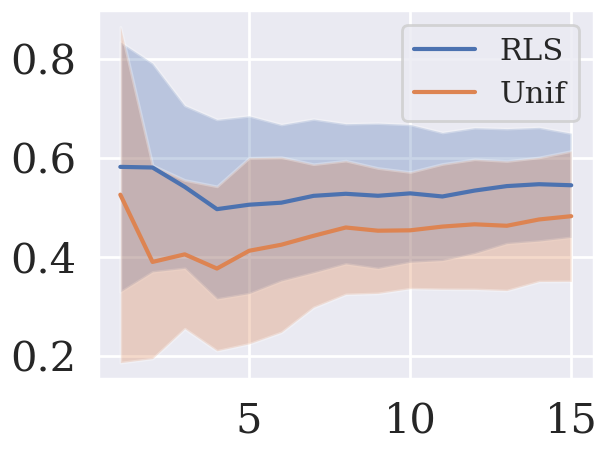}
    \end{minipage}%
    \caption{ \textsc{Ring}: number of modes (first) and sample quality (second). \textsc{Grid}: number of modes (third) and sample quality (fourth). The $x$-axis refers to the number of generators of MwuGAN. Sample quality is assessed by counting the number of modes within $3\sigma$ of each mode. The blue (resp. red) curve indicates the results obtained for RLS (resp. unif.) sampling with MwuGAN. }
	\label{fig:nmbGen_Synt}
\end{figure}

\noindent \textbf{Impact of the number of generators in MwuGAN.}
RLSs can also be used as initial starting weights in MwuGAN. The RLSs are constructed with a Gaussian kernel with bandwidth $\sigma= 0.15$ and regularization $\gamma = 10^{-3}$. Both methods contain a mixture of $15$ GANs. This approach is compared with the classical initialization with uniform weights. In Figure~\ref{fig:nmbGen_Synt}, a clear improvement over a uniform initialization is visible when RLSs are used as initial weights. 
%}

\FloatBarrier
\subsection{Unbalanced MNIST}
We describe here an ablation study for different ridge regularization parameters $\gamma>0$ and size of dimension reduction $k$ on the \textsc{unbalanced $012$-MNIST} and \textsc{unbalanced MNIST} dataset. Moreover, the use of sketching the fixed explicit feature map is analyzed. From these results is concluded that the non-linear UMAP procedure leads to better mode coverage. The results are given in Tables~\ref{tab:012MNIST_ablation} and \ref{tab:FullMNIST_ablation}. 

\begin{table}[h]
    \caption{Ablation study  over the regularization parameter $\gamma$ and dimension $k$ on the \textsc{unbalanced 012-MNIST} dataset for RLS BuresGAN with a  discriminator feature map (Discr.) with sketching and an explicit feature map obtained from the next-to-last layer of a pre-trained classifier (Class.) with a UMAP dimensionality reduction and sketched dimensionality reduction. Minority modes are highlighted in black in the first row. The best performance is achieved using the Discr. sketched and Class. UMAP. feature maps with regularization $10^{-4}$ and dimension $k=25$.}
    \label{tab:012MNIST_ablation}
\begin{center}

    \begin{tabular}{l c c c c c c c}
        \toprule
      Feature map   & Dim.  & $\quad \gamma \quad$   & Mode 1 & Mode 2 & \textbf{Mode 3}  & KL \\
   \cmidrule(lr){1-1}\cmidrule(lr){2-2} \cmidrule(lr){3-3}\cmidrule(lr){4-4}\cmidrule(lr){5-5} \cmidrule(lr){6-6} \cmidrule(lr){7-7}
\multirow{6}{*}{Discr. Sketched}   & \multirow{3}{*}{10} &  $10^{-2}$ & $5561(381)$ & $2756(300)$ & $1409(345)$ & $0.16(0.05)$   \\
    &   &   $10^{-3}$ & $5568(298)$ & $2769(178)$ & $1386(198)$ & $0.15(0.03)$ \\
    &   &   $10^{-4}$ & $5655(279)$ & $2812(135)$ & $1218(253)$ & $0.17(0.04)$  \\ \cmidrule(lr){2-7}
& \multirow{3}{*}{25} &  $10^{-2}$ & $5814(379)$ & $2405(225)$ & $1501(247)$ & $0.17(0.04)$   \\ 
    &   &   $10^{-3}$ & $5715(312)$ & $2477(139)$ & $1415(310)$ & $0.17(0.04)$  \\
    &   &   $10^{-4}$ & $5748(172)$ & $2416(268)$ & $1566(293)$ & $0.16(0.02)$   \\ \midrule
\multirow{6}{*}{Class. UMAP}  & \multirow{3}{*}{10} &  $10^{-2}$ & $3711(173)$ & $5361(103)$ & $751(161)$ & $0.18(0.03)$  \\
    &   &   $10^{-3}$ & $3813(195)$ &  $5425(208)$ &  $644(67)$ & $0.19(0.01)$ \\
    &   &   $10^{-4}$ & $3713(221)$ & $5431(226)$ & $712(152)$ & $0.18(0.03)$   \\ \cmidrule(lr){2-7}
& \multirow{3}{*}{25} &  $10^{-2}$ & $4221(104)$ & $4958(137)$ & $672(109)$ & $0.18(0.02)$ \\ 
   &   &   $10^{-3}$ & $4179(111)$ & $4876(148)$ & $802(105)$ & $0.16(0.01)$  \\
    &   &   $10^{-4}$ & $3414(161)$ & $4862(134)$ & $1461(183)$ & $0.08(0.01)$ \\ \midrule
\multirow{6}{*}{Class. Sketched}  & \multirow{3}{*}{10}&  $10^{-2}$ & $5865(122)$ & $3389(79)$ &  $605(111)$ & $0.25(0.03)$\\
    &   &   $10^{-3}$ & $5715(241)$ & $3456(216)$ & $683(161)$ & $0.31(0.01)$  \\
    &   &   $10^{-4}$ & $5952(309)$ & $3403(206)$ &  $513(135)$ &  $0.27(0.04)$  \\ \cmidrule(lr){2-7}
& \multirow{3}{*}{25} &  $10^{-2}$ & $5998(193)$ & $3130(105)$ & $672(119)$ & $0.25(0.03)$  \\ 
   &   &   $10^{-3}$ & $5957(100)$ & $3127(136)$ & $775(95)$ & $0.23(0.01)$ \\
    &   &   $10^{-4}$ & $6009(157)$ & $3138(142)$ &  $698(121)$ & $0.32(0.01)$  \\ 
         \bottomrule
    \end{tabular}

\end{center}
\end{table}

\begin{table}[h]
    \caption{Ablation study  over the regularization parameter $\gamma$ and dimension $k$ on the \textsc{unbalanced MNIST} dataset for RLS BuresGAN with a  discriminator feature map (Discr.) with sketching and an explicit feature map obtained from the next-to-last layer of a pre-trained classifier (Class.) with a UMAP dimensionality reduction and sketched dimensionality reduction. Minority modes are highlighted in black in the first row. The best performance is achieved using the Class. UMAP. feature map.}
    \label{tab:FullMNIST_ablation}
\begin{center}
\rotatebox{90}{
\resizebox{0.82\textheight}{!}{
    \begin{tabular}{l c c c c c c c c c c c c c c}
        \toprule
      Feature map   & Dim.  & $\quad \gamma \quad$   & \textbf{Mode 1} & \textbf{Mode 2} & \textbf{Mode 3}  & \textbf{Mode 4}  & \textbf{Mode 5}  & Mode 6 & Mode 7 & Mode 8 & Mode 9 & Mode 10 & KL \\
   \cmidrule(lr){1-1}\cmidrule(lr){2-2} \cmidrule(lr){3-3}\cmidrule(lr){4-4}\cmidrule(lr){5-5} \cmidrule(lr){6-6} \cmidrule(lr){7-7} \cmidrule(lr){8-8} \cmidrule(lr){9-9} \cmidrule(lr){10-10} \cmidrule(lr){11-11} \cmidrule(lr){12-12} \cmidrule(lr){13-13} \cmidrule(lr){14-14}
\multirow{6}{*}{Discr. Sketched}   & \multirow{3}{*}{10} &  $10^{-2}$ & $254(77)$ & $170(30)$ & $222(40)$ & $227(46)$ & $212(62)$ & $2061(94)$ & $1871(134)$ & $1598(71)$ & $1853(83)$ & $1532(96)$  & $0.38(0.02)$ \\
    &   &   $10^{-3}$ & $214(38)$ & $172(123)$ & $239(35)$ & $223(99)$ & $230(54)$ & $2081(91)$ & $1832(32)$ & $1611(100)$ & $1898(44)$ & $1501(123)$ & $0.39(0.02)$ \\
    &   &   $10^{-4}$ & $235(62)$ & $183(141)$ & $264(44)$ & $255(109)$ & $219(54)$ & $2012(97)$ & $1839(53)$ & $1608(155)$ & $1841(55)$ & $1545(115)$ & $0.37(0.02)$ \\ \cmidrule(lr){2-14}
& \multirow{3}{*}{25} &  $10^{-2}$ &  $236(56)$ & $191(145)$ & $286(43)$ & $231(72)$ & $219(60)$ & $2066(113)$ & $1765(48)$ & $1505(81)$ & $1958(37)$ & $1543(56)$ &  $0.37(0.02)$  \\ 
    &   &   $10^{-3}$ &  $230(30)$ & $172(129)$ & $267(31)$ & $248(89)$ & $274(68)$ & $2046(100)$ & $1780(42)$ & $1546(122)$ & $1899(73)$ & $1538(135)$ & $0.37(0.02)$ \\
    &   &   $10^{-4}$ &  $224(49)$ & $147(179)$ & $255(27)$ & $223(52)$ & $197(53)$ & $2032(116)$ & $1790(49)$ & $1554(119)$ & $1962(55)$ & $1617(143)$ & $0.40(0.02)$  \\ \midrule
\multirow{6}{*}{Class. UMAP}  & \multirow{3}{*}{10} &  $10^{-2}$ & $885(79)$ & $662(143)$ & $359(229)$ & $814(97)$ & $578(66)$ & $1248(152)$ & $1137(152)$ & $1847(154)$ & $1095(98)$ & $1374(112)$ & $0.10(0.01)$  \\
    &   &   $10^{-3}$ & $892(137)$ & $674(90)$ & $370(227)$ & $803(108)$ & $605(75)$ & $1187(189)$ & $1123(103)$ & $1784(111)$ & $1110(95)$ & $1452(189)$ & $0.09(0.02)$\\
    &   &   $10^{-4}$ & $875(112)$ & $663(122)$ & $360(198)$ & $831(59)$ & $615(82)$ & $1311(178)$ & $1137(116)$ & $1743(79)$ & $1069(67)$ & $1396(100)$ & $0.09(0.01)$  \\ \cmidrule(lr){2-14}
& \multirow{3}{*}{25} &  $10^{-2}$ & $556(60)$ & $525(82)$ & $352(59)$ & $542(71)$ & $473(56)$ & $1471(99)$ & $1449(87)$ & $1915(101)$ & $1327(66)$ & $1392(80)$ & $0.16(0.01)$ \\ 
   &   &   $10^{-3}$ & $495(58)$ & $473(114)$ & $385(50)$ & $546(83)$ & $463(50)$ & $1450(120)$ & $1452(69)$ & $1964(110)$ & $1290(75)$ & $1483(81)$ & $0.17(0.02)$\\
    &   &   $10^{-4}$ & $499(73)$ & $443(103)$ & $358(48)$ & $513(104)$ & $450(45)$ & $1491(148)$ & $1468(51)$ & $1957(119)$ & $1329(43)$ & $1493(145)$ & $0.18(0.01)$ \\ \midrule
\multirow{6}{*}{Class. Sketched}  & \multirow{3}{*}{10}&  $10^{-2}$ & $218(72)$ & $188(171)$ & $165(31)$ & $261(114)$ & $246(41)$ & $2019(166)$ & $1726(48)$ & $1859(204)$ & $1683(63)$ & $1637(178)$ & $0.38(0.02)$ \\
    &   &   $10^{-3}$ & $189(45)$ & $194(238)$ & $184(88)$ & $245(232)$ & $221(44)$ & $2147(176)$ & $1712(38)$ & $1824(182)$ & $1666(61)$ & $1618(177)$ & $0.4(0.03)$ \\
    &   &   $10^{-4}$ & $239(45)$ & $127(200)$ & $211(27)$ & $232(133)$ & $214(65)$ & $2127(113)$ & $1783(48)$ & $1804(122)$ & $1642(29)$ & $1622(142)$ & $0.48(0.02)$ \\ \cmidrule(lr){2-14}
& \multirow{3}{*}{25} &  $10^{-2}$ & $238(30)$ & $192(205)$ & $189(35)$ & $308(84)$ & $254(39)$ & $2011(105)$ & $1699(74)$ & $1710(127)$ & $1785(54)$ & $1615(143)$ & $0.36(0.02)$\\ 
   &   &   $10^{-3}$ & $262(45)$ & $204(116)$ & $251(45)$ & $258(83)$ & $266(62)$ & $2041(152)$ & $1621(40)$ & $1774(119)$ & $1751(41)$ & $1573(134)$ & $0.35(0.02)$ \\
    &   &   $10^{-4}$ &  $246(50)$ & $220(131)$ & $199(41)$ & $312(70)$ & $306(41)$ & $2033(127)$ & $1689(38)$ & $1712(115)$ & $1724(67)$ & $1560(100)$ & $0.34(0.02)$ \\ 
         \bottomrule
    \end{tabular}
    }
    }
\end{center}
\end{table}

\FloatBarrier
\section{Timings\label{B}}
Training times with a single NVIDIA Tesla P100-SXM2-16GB @1.3 GHz GPU. Temporary evaluations and plotting during training are included in the timings, so only relative comparisons are instructive. 
\begin{table}[h]
    \caption{Total training time in seconds, averaged over 10 runs.}
    \label{tab:timings}
\begin{center}
    \begin{tabular}{l c c c c}
        \toprule
          & Ring & Grid & 012-MNIST & MNIST\\
           \cmidrule(lr){2-2}  \cmidrule(lr){3-3}  \cmidrule(lr){4-4} \cmidrule(lr){5-5} 
         GAN  & $45(1)$  & $48(1)$ & $1063(4)$ & $1074(3)$  \\
         PacGAN2   & $ 65(1)$  & $71(1)$ & $1205(4)$ & $1211(8)$  \\
         BuresGAN  & $100(1)$ & $105(2)$  & $1357(3)$ & $1373(3)$ \\ \midrule
         IwGAN & $65(1)$ & $65(1)$ & $1066(5)$ & $1080(3.3)$ \\
         IwMmdGAN & $45(1)$ & $56(1)$ &  $519(4)$ & $5126(1)$  \\ \midrule
         MwuGAN (15) & $1195(9)$  &  $1225(8)$  & $20730(314)$  & $36074(292)$  \\ 
         RLS MwuGAN (15) & $ 1198(9)$  & $ 1217(9)$  & $20823(233)$ & $35400(566)$    \\\midrule
         RLS BuresGAN Gauss. & $130(3)$  & $136(2)$ &  / & / \\ 
         RLS BuresGAN Class. & /  & / & $1615(11)$ & $2494(8)$ \\ 
         RLS BuresGAN Discr. & $ 176(4)$  & $187(1)$  & $2006(3)$ & $2004(2)$ \\ 
         \bottomrule
    \end{tabular}
\end{center}
\end{table}
\FloatBarrier
\section{Distribution of generated samples for the artificial datasets \label{C}}
\begin{figure}[h]
	\centering
    \begin{minipage}{0.24\textwidth}
        \centering
        \includegraphics[width=1\linewidth]{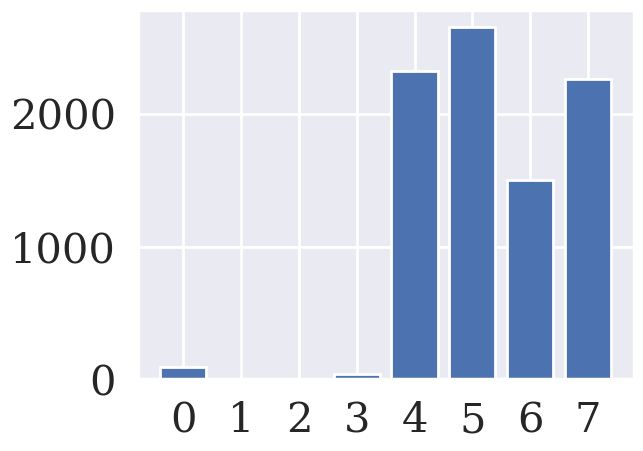}
       \text{\hspace{8mm} GAN} \vspace{3mm}
    \end{minipage}
    \begin{minipage}{.24\textwidth}
        \centering
        \includegraphics[width=1\linewidth]{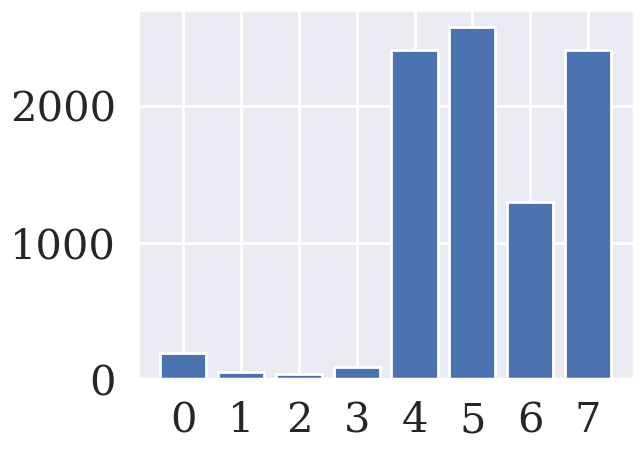}
        \text{\hspace{8mm} PacGAN2} \vspace{3mm}
    \end{minipage}%
    \begin{minipage}{0.24\textwidth}
        \centering
        \includegraphics[width=1\linewidth]{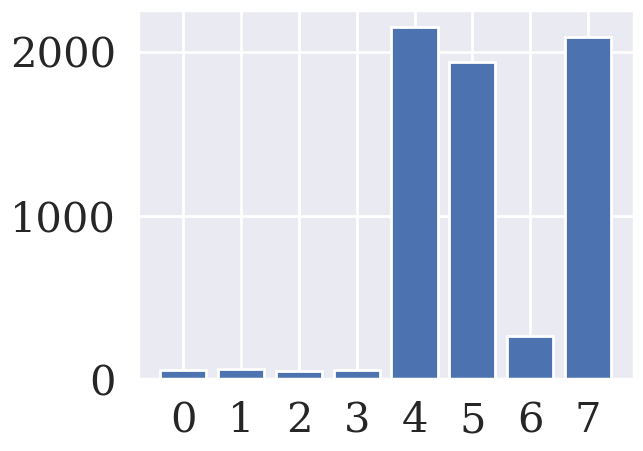}
        \text{\hspace{8mm} BuresGAN} \vspace{3mm}
    \end{minipage}
    \begin{minipage}{0.24\textwidth}
        \centering
        \includegraphics[width=1\linewidth]{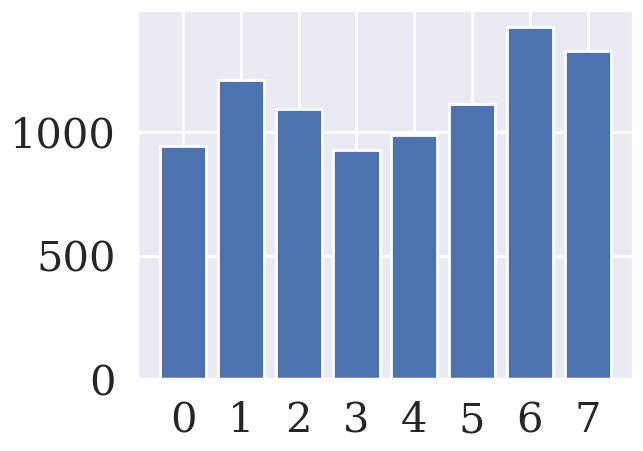}
        \text{\hspace{8mm} IwGAN} \vspace{3mm}
    \end{minipage}
    \begin{minipage}{0.24\textwidth}
        \centering
        \includegraphics[width=1\linewidth]{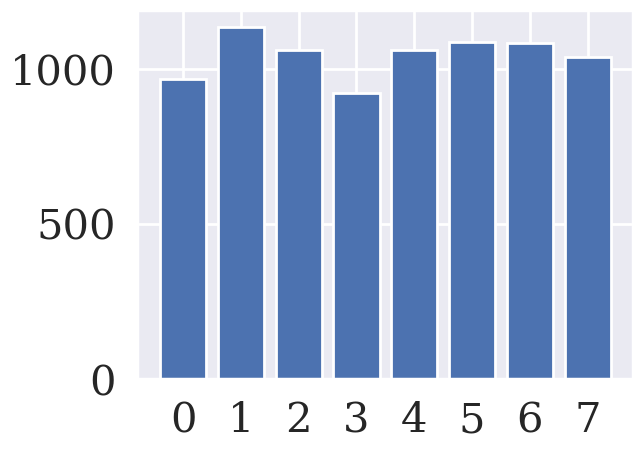}
        \text{\hspace{8mm} IwMmdGAN} \vspace{3mm}
    \end{minipage}
    \begin{minipage}{.24\textwidth}
        \centering
        \includegraphics[width=1\linewidth]{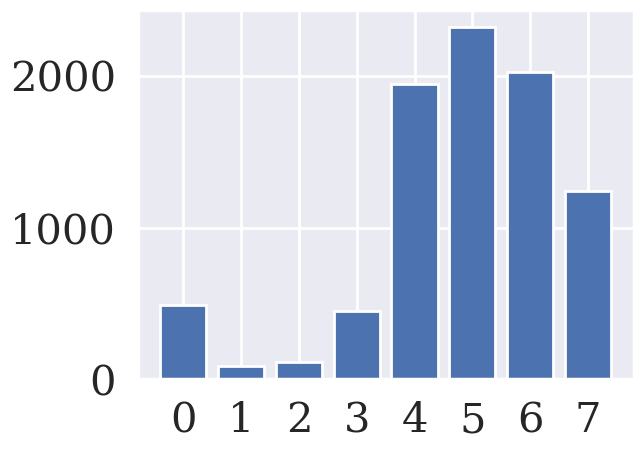}
        \text{\hspace{8mm} MwuGAN} \vspace{3mm}
    \end{minipage}
    \begin{minipage}{.24\textwidth}
        \centering
        \includegraphics[width=1\linewidth]{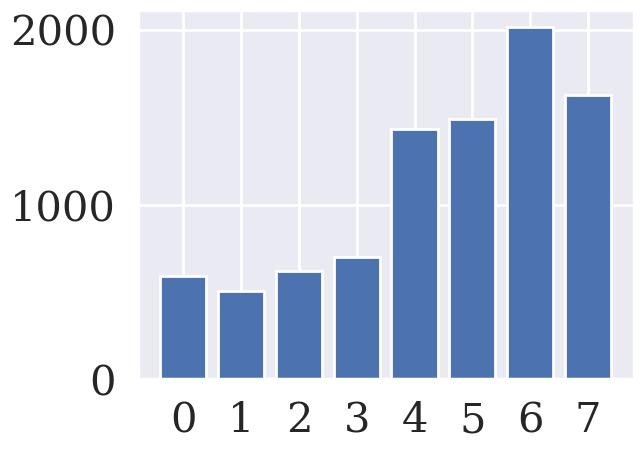}
        \text{\hspace{5mm} RLS MwuGAN} \vspace{3mm}
    \end{minipage}
    \begin{minipage}{.24\textwidth}
        \centering
        \includegraphics[width=1\linewidth]{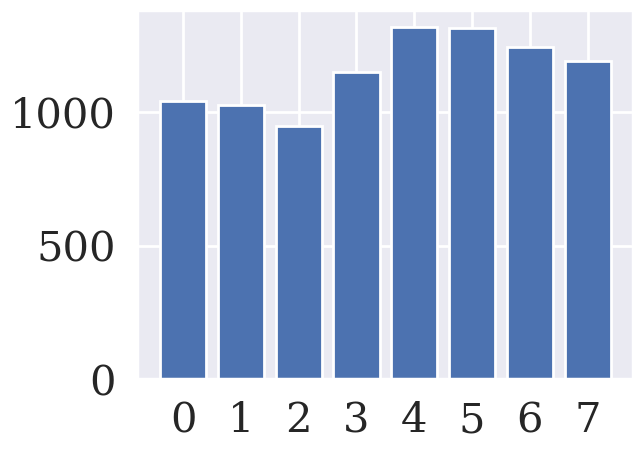}
        \text{\hspace{5mm} RLS BuresGAN} \vspace{0mm}
        \text{\hspace{5mm} Discr.} \vspace{0mm}
    \end{minipage}
    \caption{Samples per mode for a single GAN trained on the \textsc{Ring} dataset.  Only  IwMmdGAN, MwuGAN (15), RLS MwuGAN (15), and RLS BuresGAN with a discriminator feature map (Discr.) are capable of covering all the modes.}
	\label{fig:Samp_Ring_all}
\end{figure}

\begin{figure}[h]
	\centering
    \begin{minipage}{0.24\textwidth}
        \centering
        \includegraphics[width=1\linewidth]{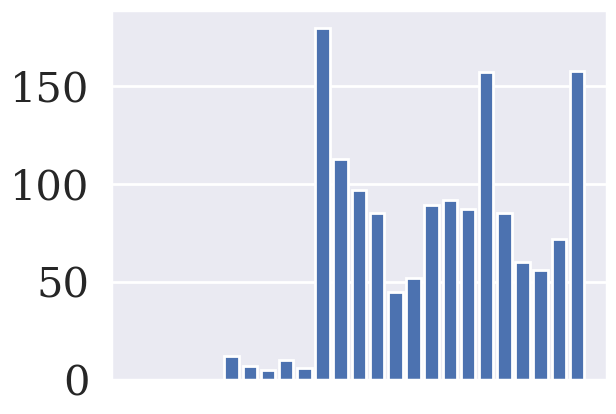}
        \text{\hspace{8mm} GAN} \vspace{3mm}
    \end{minipage}
    \begin{minipage}{.24\textwidth}
        \centering
        \includegraphics[width=1\linewidth]{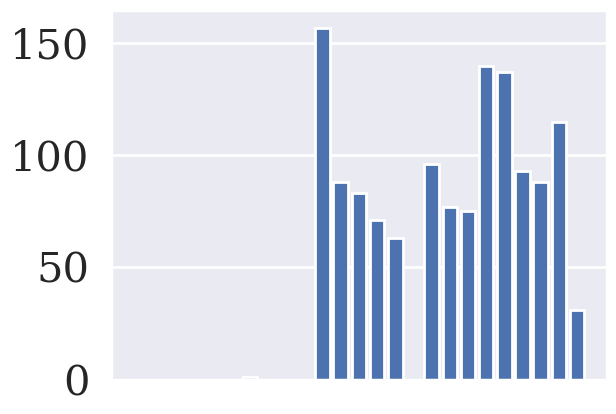}
        \text{\hspace{8mm} PacGAN2} \vspace{3mm}
    \end{minipage}%
    \begin{minipage}{0.24\textwidth}
        \centering
        \includegraphics[width=1\linewidth]{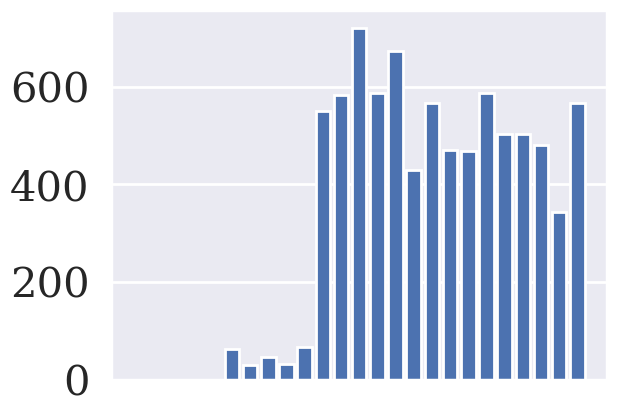}
        \text{\hspace{8mm} BuresGAN} \vspace{3mm}
    \end{minipage}
    \begin{minipage}{0.24\textwidth}
        \centering
        \includegraphics[width=1\linewidth]{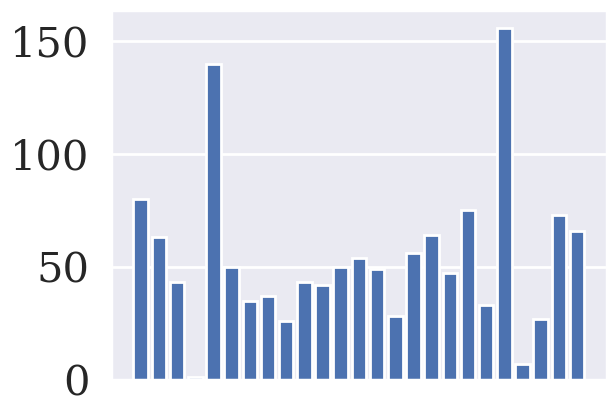}
        \text{\hspace{8mm} IwGAN} \vspace{3mm}
    \end{minipage}
    \begin{minipage}{0.24\textwidth}
        \centering
        \includegraphics[width=1\linewidth]{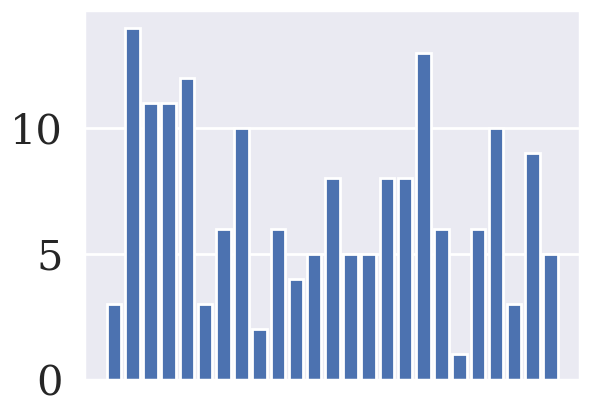}
        \text{\hspace{8mm} IwMmdGAN} \vspace{3mm}
    \end{minipage}
    \begin{minipage}{.24\textwidth}
        \centering
        \includegraphics[width=1\linewidth]{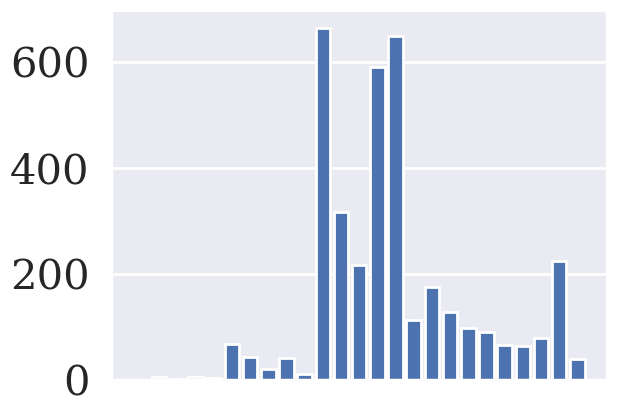}
        \text{\hspace{8mm} MwuGAN} \vspace{3mm}
    \end{minipage}%
    \begin{minipage}{.24\textwidth}
        \centering
        \includegraphics[width=1\linewidth]{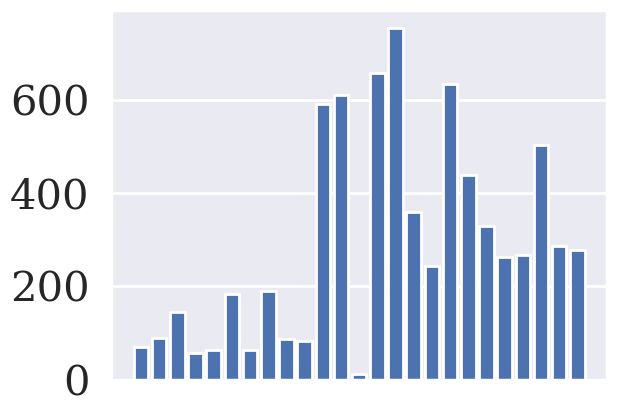}
        \text{\hspace{5mm} RLS MwuGAN} \vspace{3mm}
    \end{minipage}%
    \begin{minipage}{.24\textwidth}
        \centering
        \includegraphics[width=1\linewidth]{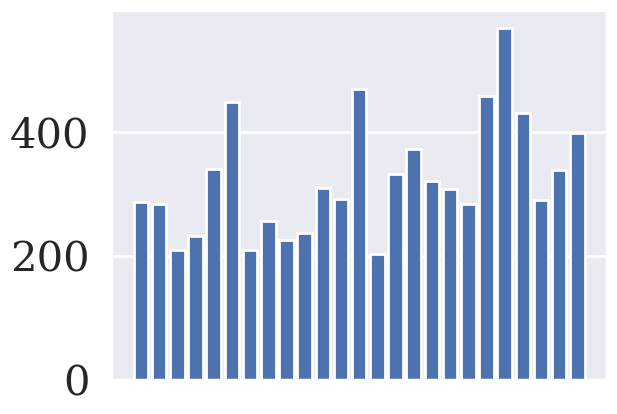}
        \text{\hspace{5mm} RLS BuresGAN} \vspace{0mm}
        \text{\hspace{5mm} Discr.} \vspace{0mm}
    \end{minipage}%
    \caption{Samples per mode for a single GAN trained on the \textsc{Grid} dataset. Only the proposed  RLS BuresGAN with a discriminator feature map (Discr.) is capable of covering all the modes.}
	\label{fig:Samp_Grid_all}
\end{figure}

The number of samples per mode for all GANs in the \textsc{Ring} experiments are given in Figure~\ref{fig:Samp_Ring_all}. The number of samples per mode for all GANs in the \textsc{Grid} experiments are given in Figure~\ref{fig:Samp_Grid_all}.

\FloatBarrier
\section{Architectures, algorithms, and settings\label{D}}

\noindent\textbf{Settings experimental section.} Unless specified otherwise, the models are trained for $30$k iterations with a batch size of $64$, by using the Adam optimizer with $\beta_1 = 0.5$, $\beta_2=0.999$ and learning rate $10^{-3}$ for both the generator and discriminator. The dimensionality of the generator latent space $\ell$ is equal to $100$. The results of the experimental section always refer to the performance achieved at the end of the training.  All the images are scaled in between -1 and 1 before running the algorithms. The hyperparameters of the competing methods are chosen as suggested in the authors' reference implementations. For MwuGAN, we take $\delta = 0.25$ and run for an increasing number of generators (displayed in brackets). In all the simulations, PacGAN always uses $2$ as packing number. Note that IwMmdGAN and IwGAN require full knowledge of the Radon-Nikodym derivative $M$, where we want to generate data from a target distribution $p$ but only have access to representative samples from a modified distribution $Mp$. The method is thus not an unsupervised method and tackles a slightly different problem. However, the method was included for completeness. \\ 

\begin{algorithm}[h]
 \caption{RLS BuresGAN with a fixed feature map.}
\textbf{Input:} $ \{x_{i}\}_{i=1}^n$, regularization $\gamma$, dimension reduction size $k$, discriminator $D$, generator $G$, feature map $ \varphi({\cdot}) \in \mathbb{R}^m$ - explicit \emph{or} implicit via kernels $ k(\cdot, \cdot)$  \\
\textbf{Calculate ridge leverage scores.}
	\begin{algorithmic}[1]
			\If{$\varphi ({\cdot})$ = Implicit}
			\State Calculate the RLSs for all points: $\ell_{i} = \left(K(K+ n\gamma \mathbb{I})^{-1}\right)_{i i}$.
			\ElsIf{$\varphi({\cdot})$ = Explicit}
			\If{dimension reduction}
			\State $\varphi(x_i) \leftarrow$ UMAP of size $k$ calculated on the full projected dataset $ \{\varphi(x_i)\}_{i=1}^n$.
			\Else
			\State $k \leftarrow m$.
			\EndIf
			\If{$n > k$} 
			\State Calculate the RLSs for all points: $\ell_{i} = \varphi(x_i)^\top(C + n\gamma \mathbb{I})^{-1}\varphi(x_i)$.
			\Else
			\State Calculate the RLSs for all points: $\ell_{i} = \left(K(K+ n\gamma \mathbb{I})^{-1}\right)_{i i}$.
			\EndIf
			\EndIf
\end{algorithmic}
\textbf{Train BuresGAN.}
	\begin{algorithmic}[1]
	        \While{not converged}
			\State Sample a real and fake batch with probability $x_i \sim \ell_{i}$.
			\State Update the generator $G$ by minimizing $V_G +  \mathcal{B}(\hat{C}_{r}, \hat{C}_{g})^{2}$
			\State Update the discriminator $D$ by maximizing $- V_D$
			\EndWhile
\end{algorithmic}	
\end{algorithm}

\begin{algorithm}[h]
 \caption{RLS BuresGAN with a discriminator-based explicit feature map.}
\textbf{Input:} $ \{x_{i}\}_{i=1}^n$, regularization $\gamma$, dimension reduction size $k$, mini-batch size $s$, discriminator $D$ with discriminator feature map $\varphi_D(\cdot) \in \mathbb{R}^m$, generator $G$. \\
\textbf{Train BuresGAN with adaptive RLSs.}
	\begin{algorithmic}[1]
	        \While{not converged}
	        \State Sample a mini-batch $\hat{x}$ uniform at random of size $b = 20s$.
	        \If{dimension reduction}
			\State $\varphi(\hat{x}_i) \leftarrow S^\top\varphi_D(\hat{x}_i) \in \mathbb{R}^{k}$, with sketching matrix $S= A/\sqrt{k} \in \mathbb{R}^{m \times k}$.
			\Else
			\State $\varphi(\hat{x}_i) \leftarrow \varphi_D(\hat{x}_i)$, $k \leftarrow m$.
			\EndIf
			\If{$b > k$} 
			\State Calculate the RLSs for all points: $\ell_{i} = \varphi(x_i)^\top(C + n\gamma \mathbb{I})^{-1}\varphi(x_i)$.
			\Else
			\State Calculate the RLSs for all points: $\ell_{i} = \left(K(K+ n\gamma \mathbb{I})^{-1}\right)_{i i}$.
			\EndIf
			\State Sample a real and fake batch of size $s$ with probability $\hat{x}_i \sim \ell_{i}$.
			\State Update the generator $G$ by minimizing $V_G +  \mathcal{B}(\hat{C}_{r}, \hat{C}_{g})^{2}$
			\State Update the discriminator $D$ by maximizing $- V_D$
			\EndWhile
\end{algorithmic}
\end{algorithm}

\begin{table}[h]
    \caption{The generator (left) and discriminator (right) architectures for the synthetic examples.}
    \label{tab:gen_dis_architecture_synthetic}
\begin{center}

    \begin{tabular}{l c c c }
        \toprule
          Layer & Output & Activation  \\%& Time \\ 
          \cmidrule(lr){1-3}
         Input  & 25  & -  \\
         Dense  & 128  & tanh  \\
         Dense  & 128  &  tanh \\
         Dense   &  2  &  - \\
         \bottomrule
    \end{tabular} \quad \quad
    \begin{tabular}{l c c c }
        \toprule
          Layer & Output & Activation  \\%& Time \\ 
          \cmidrule(lr){1-3}
         Input  & 2  & -  \\
         Dense  & 128  & tanh  \\
         Dense  & 128  &  tanh \\
         Dense   &  1  &  - \\
         \bottomrule
    \end{tabular}

\end{center}
\end{table}

\begin{table}[h]
    \caption{The generator (left) and discriminator (right) architectures for the \textsc{unbalanced $012$-MNIST} and the \textsc{unbalanced MNIST} experiments. BN indicates if batch normalization is used.}
    \label{tab:MNISTparameters}
\begin{center}

    \begin{tabular}{l c c c c}
        \toprule
          Layer & Output & Activation & BN \\%& Time \\ 
          \cmidrule(lr){1-4}
         Input  & 100  & -  & -\\
         Dense  & 12544  & ReLU  & Yes\\
         Reshape   &  7, 7, 256  &  - & -\\
         Conv'   &  7, 7, 128  &  ReLU & Yes\\
         Conv'   &  14, 14, 64  &  ReLU & Yes\\
         Conv'   &  28, 28, 1  &  ReLU & Yes\\
         \bottomrule
    \end{tabular} \quad \quad
      \begin{tabular}{l c c c c}
        \toprule
          Layer & Output & Activation & BN \\%& Time \\ 
          \cmidrule(lr){1-4}
         Input  & 28, 28, 1  & -  & -\\
         Conv  & 14, 14, 64  & Leaky ReLU  & No \\
         Conv  & 7, 7, 128  &  Leaky ReLU & Yes\\
         Conv  & 4, 4, 256  &  Leaky ReLU & Yes\\
         Conv  & 2, 2, 512  &  Leaky ReLU & Yes\\
         Flatten  & - &  - & - \\
         Dense  & 1  &  - & -\\
         \bottomrule
    \end{tabular}

\end{center}
\end{table}

\begin{table}[t]
    \caption{The CNN architecture of the classifier used during the evaluation of the MNIST experiments. Dropout with a rate of 0.5 is used before the final dense layer.}
    \label{tab:MNISTclassifier}
\begin{center}

    \begin{tabular}{l c c}
        \toprule
          Layer & Output & Activation \\%& Time \\ 
          \cmidrule(lr){1-3}
         Input  & 28, 28,1  & -  \\
         Conv   &  24, 24, 32  &  ReLU \\
         MaxPool   &  12, 12, 32  &  - \\
         Conv   &  8, 8, 64  &  ReLU \\
         MaxPool   &  4, 4, 64  &  - \\
         Flatten   &  -  &  - \\
         Dense   &  1024  &  ReLU \\
         Dense   &  10  &  - \\
         \bottomrule
    \end{tabular} \quad \quad

\end{center}
\end{table}

\end{document}